\documentclass[11pt,a4paper]{article}

\RequirePackage[utf8]{inputenc}
\usepackage{graphicx}
	\newcommand {\be}{\begin{equation}}
	\newcommand {\ee}{\end{equation}}
	\newcommand {\bq}{\begin{eqnarray}}
	\newcommand {\eq}{\end{eqnarray}}
	\renewcommand {\l}[1]{\lambda_{#1}}
\begin{document}

\title{Virtual Image Correlation uncertainty}
\author{M. L. M. François
\footnote{Marc Louis Maurice François, laboratory GeM, UMR CNRS 6183,
CNRS---Université de Nantes --- \'Ecole Centrale Nantes, 2, rue de la Houssinière BP 92208, 44322 Nantes Cedex 3, France
Tel.: +33-2-51125521
marc.francois@univ-nantes.fr
}}
\date{\today}
\maketitle

\begin{abstract}
The Virtual Image Correlation method applies for the measurement of silhouettes boundaries with sub-pixel precision.
It consists in a correlation between the image of interest and a virtual image based on a parametrized curve.
Thanks to a new formulation, it is shown that the method is exact in 1D, insensitive to local curvature and to contrast variation, and that the  bias induced by luminance variation can be easily corrected.
Optimal value of the virtual image width, the sole parameter of the method, and optimal numerical settings are established.
An estimator is proposed to assess the relevance of the user-chosen curve to describe the contour with a sub-pixel precision.
Analytical formulas are given for the measurement uncertainty in both cases of noiseless and noisy images and their prediction is successfully compared to numerical tests.
\end{abstract}

\noindent
The software "Funambule" associated with this publication is available at\\\texttt{https://zenodo.org/record/3862248}, DOI : 10.5281/zenodo.3862248\\
or, for latest version, \\
\texttt{https://github.com/marc-l-m-francois/Funambule/releases}

%Your text comes here. Separate text sections with
%\section{Section title}
%\label{sec:1}
%Text with citations \cite{RefB} and \cite{RefJ}.
%\subsection{Subsection title}
%\label{sec:2}
%as required. Don't forget to give each section
%and subsection a unique label (see Sect.~\ref{sec:1}).
%\paragraph{Paragraph headings} Use paragraph headings as needed.
%\begin{equation}
%\end{equation}
%% For one-column wide figures use
%\begin{figure}
%% Use the relevant command to insert your figure file.
%% For example, with the graphicx package use
%%  \includegraphics{example.eps}
%\end{figure}
%%
%% For two-column wide figures use
%\begin{figure*}
%% Use the relevant command to insert your figure file.
%% For example, with the graphicx package use
%%  \includegraphics[width=0.75\textwidth]{example.eps}
%\end{figure*}
%\begin{table}
%\begin{tabular}{lll}
%\hline\noalign{\smallskip}
%\end{tabular}
%\end{table}

%---------------------------------------------------------------
\section{Introduction}
\label{intro}
%---------------------------------------------------------------

The Virtual Image Correlation (VIC) originates from the global form of the Digital Image Correlation (DIC) method \cite{Hild2006,Hild2006b}.
However, in the VIC, the second image is an elementary and unitary virtual one which mimics the white to black gradient of the boundary and whose shape is defined from a parametrized curve.
At convergence, when virtual and physical images are close as possible, the curve shape represents a measurement of the contour.\\

The very first version of the VIC was dedicated to open contour measurement \cite{francois10amm}.
Its extension to silhouette measurement followed \cite{francois11epj,CONFfrancois11bes_a} then a numerically efficient version benefitting of close DIC developments \cite{francois13exm}.
Further work concerned various application of the method for the mechanical testings \cite{CONFSEM15,bloch2015,CONFRILEM16} or in medicine \cite{Jiang2015}.
The major interest of the VIC is its precision, in some case better than $10^{-3}$ pixels \cite{francois11epj}.
However, the present article was motivated by a need for more objective evaluation of the uncertainty, with predictive formulas.\\

In Sec.~\ref{sec:1} is shown a slightly modified version of the method, in which the mean square distance between virtual and physical images is calculated in the frame of the virtual image.
This gives both a slightly better precision and much simpler equations.\\

Section~\ref{sec:2} is dedicated to the quantification of uncertainties. 
It begins by establishing a set of simplified equations, which are used at first to prove that the method is theoretically exact in 1D.
The VIC requires the chosen curve family to be able to fit the contour of interest.
Aiming sub-pixel precision, the simple observation of the obtained curve superposed to the silhouette is not sufficient to check this point.
A signed distance is proposed, which consists in a local measurement of the silhouette in the virtual image frame.
Its graph emphasis the local accuracy of the identification and its spectral analysis informs about the relevance of the chosen curve to depict the contour.
The image discretization is an inevitable cause of uncertainty that leads to the ultimate accuracy of the method.
An empirical law, deducted from statistics on numerical tests, is proposed to assess it.
Effect of imperfect brightness and contrast are studied analytically.
It is shown that contrast has no effect on the precision but that brightness induces a bias which can be suppressed by a linear correction of the image.
Then, the measurement uncertainty due to image noise is quantified by a simple analytical formula.
A simple graph summarizes the expected accuracy as a function of the image noise and the number of parameters of the curve.
\\

Section~\ref{sec:3} validates the proposed expressions of uncertainties, through statistics on old \cite{francois11epj} and new synthetic tests.
In addition to the comparisons already made (in \cite{francois11epj}) with the Fast Marching Algorithm \cite{Sethian_98} and the Steger's method \cite{steger1998}, new comparisons are made here with the active contour method \cite{Lankton2019} and the recent method of Trujilo-Pino \cite{Trujillo2013} which are both known for their sub-pixel precision.
Tests on noisy images also emphasize the robustness of the VIC.
For all synthetic images used in this article, the grey levels of the transition pixels (through which the edge passes) are calculated from the ratio of the white (background) and black (silhouette) surfaces seen by the pixel.

%---------------------------------------------------------------
\section{The VIC method}\label{sec:1}
%---------------------------------------------------------------

\begin{figure}[htbp]
	\begin{center}
		\includegraphics[scale=0.333]{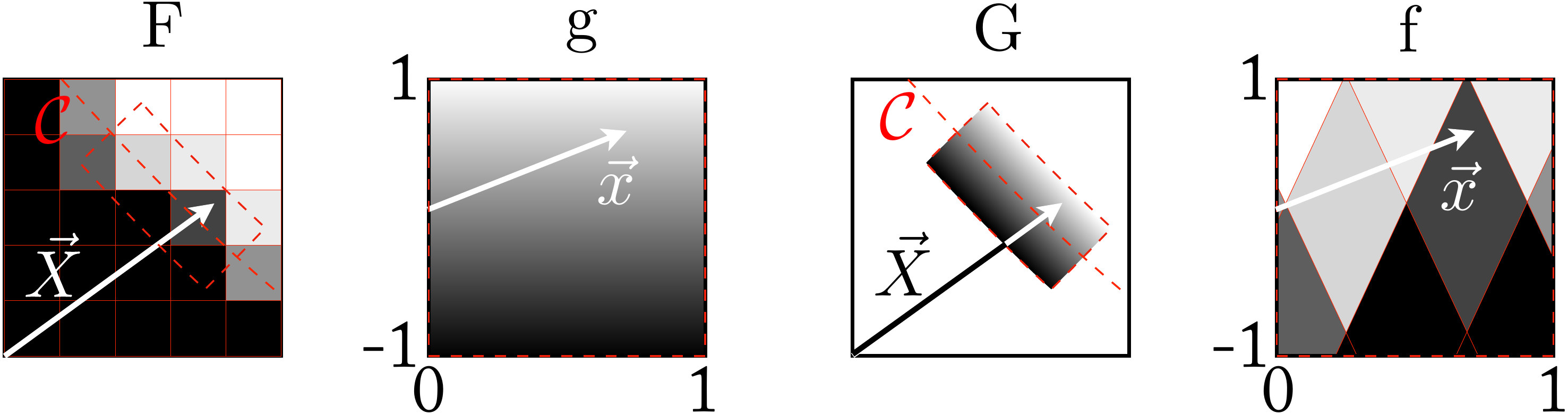}
		\caption{Sketch of the VIC method}
		\label{fig:dicvic}
	\end{center}
\end{figure}
\begin{figure}[htbp]
	\begin{center}
	\includegraphics[scale=0.5]{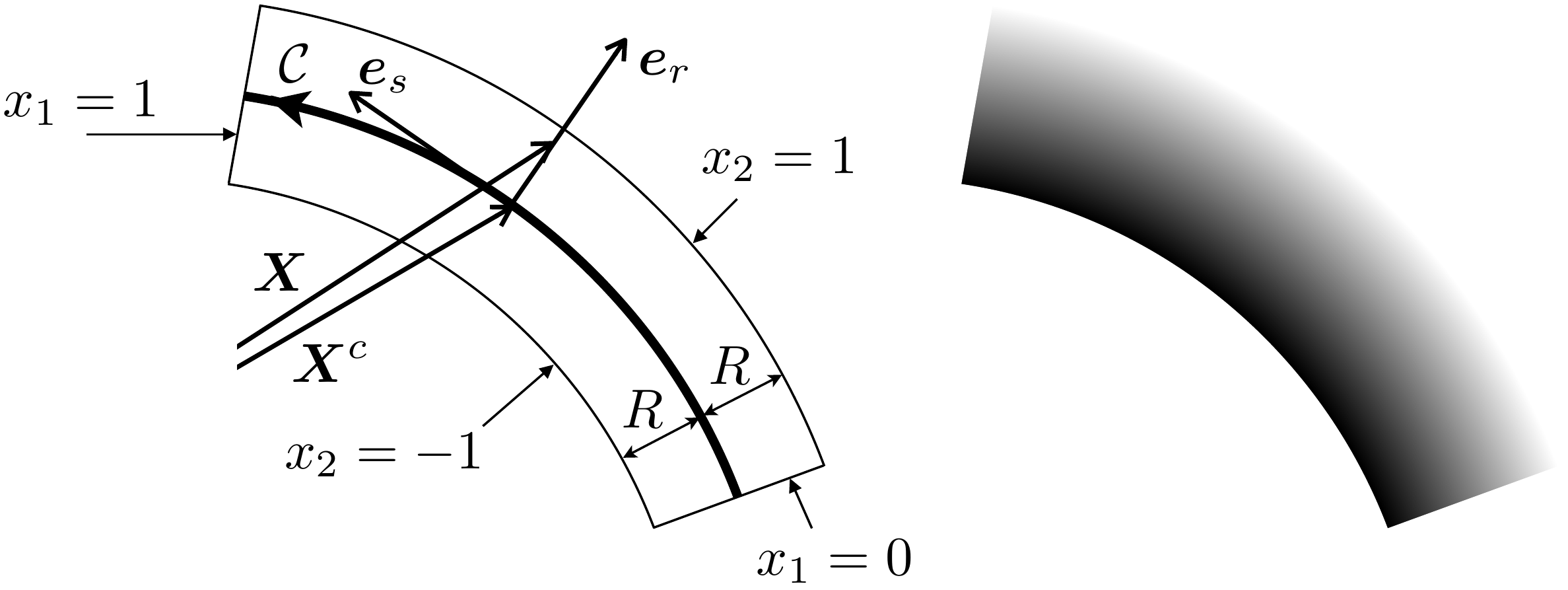}
	\caption{Virtual image geometry (left) and grey levels (right)}
	\label{fig:imG}
	\end{center}
\end{figure}

The silhouette of interest in image F is measured by finding the best coincidence between F and a virtual image G based on a parametrized curve $\mathcal C$ (see Fig.~\ref{fig:dicvic}).
Image F has grey levels $F(\vec X)$ where $\vec X$ is the position vector of components $(X_1,X_2)$ associated to the pixel frame.
The virtual image G of gray levels $G(\vec X)$ is a deformation of an elementary image g of grey levels $g(\vec x)$ such as:
\bq
	G(\vec{X}) &=& g(\vec{x})\label{eq:Gg}\\
	g(\vec x) &=& \frac{1+x_2}{2}\label{eq:gdexsil}
\eq 
where the position vector $\vec x$ has components $x_1\in[0,1]$ and $x_2\in[-1,1]$.
The linear evolution of the gray level $g(\vec x)$ is chosen in order G to be roughly similar to the gray level evolution across the boundary in F.
A current point $\vec X$ of G is defined from the user-chosen parametric curve $\mathcal C$ of current point $\vec X^c$ (see Fig.~\ref{fig:imG}):
\be
	\vec{X}(x_1,x_2,\l p)=\vec X^c(x_1,\l p)+ Rx_2\, \vec{e}_r(x_1,\l p) \label{eq:revmap},
\ee
where $x_1$ is used as the curve parameter, the $\l p$ are the (researched) shape parameters and $(\vec e_s,\vec e_r)$ are respectively the unitary tangent and normal vectors to the curve:
\bq
	\vec{e}_s &=& \frac{\partial \vec X^c}{\partial x_1}\left\|\frac{\partial \vec X^c}{\partial x_1}\right\|^{-1} \label{def:es}\\
	\vec{e}_r &=& \vec{e}_s\times\vec{e}_z	\label{def:er}
\eq
where $\times$ denotes the cross product and $\vec{e}_z$ is the unitary vector normal to the plane. 
Above definition guarantees that $\vec e_r$ points uniformly outside any closed curve orientated positively.\\
%In Eq.~(\ref{eq:revmap}), $R x_2$ defines the distance from the current point $\vec X$ to $\vec X^c$, \emph{i.e.} from $\vec X$ to the curve $\mathcal C$ in the frame $(X_1,X_2)$.\\

The goal of the method is to find the shape parameters $\l p$ of $\mathcal C$ for which F and G are in best coincidence.
As for some DIC methods \cite{Hild2006}, the mean square difference between the two images is minimized:
\bq
	\Psi &=& \frac{\int\int (F(\vec X)-G(\vec X))^2 d X_1 d X_2}{\int\int d X_1 d X_2}\label{eq:grandpsi}	
\eq
This expression was used in the very first version of the VIC \cite{francois10amm,francois11epj} in which the surface area $2RL$ (the denominator) was constant.
Neglecting the surface variation allows the use of numerically efficient DIC algorithms \cite{francois13exm} but with a loss of accuracy.
Strictly speaking, the length $L$ of the curve, so the area, is not constant. 
Furthermore, the differential surface element $d X_1 d X_2$ depends, by Eq.~(\ref{eq:dXdx}), upon the curvature and neglecting it creates a slight but unwanted line tension effect.
The proposed minimization function is expressed in the frame of the virtual image:
\bq
	\psi &=& \frac{\int\int (f(\vec x)-g(\vec x))^2 d x_1 d x_2}{\int\int d x_1 d x_2},\label{eq:petitpsi}%\\
%	f(\vec{x}) &=& F(\vec{X})\label{eq:Ff}
\eq
in which $f(\vec{x}) = F(\vec{X})$.
The denominator represents the constant surface area of g (of value 2) and the differential surface element $d x_1 d x_2$ is independent of the curvature.
%As example, any curve on a white background tends, when minimizing $\Psi$, to shrink in order to minimize the black to white ratio of G.
%On the contrary, no motion occurs when minimizing $\psi$.\\
%Such absence of line tension leads to better precision from previous methods.\\
%In detail the minimization function is:
%
%\be
%	2\psi(\l p)
%	= 
%	\int_{-1}^1\int_0^1\left(F(\vec X(x_1,x_2,\l p)) - g(x_2)\right)^2 dx_1 dx_2\label{eq:psi}
%\ee
%%
The minimization of $\Psi$ with respect to $\l p$ is achieved by using a Newton scheme, solving iteratively the $N\times N$ linear system:
%looking for the stationary point ${\partial \psi}/{\partial \l p}=0$ by an iterative resolution up to convergence of the following $N\times N$ linear system:
%
\be
	\frac{\partial^2 \psi}{\partial \l p\partial \l q}\Delta \l q = -\frac{\partial \psi}{\partial \l p}
	\label{eq:newton}
\ee
where $\Delta \l q$ is the corrector of the current values of the $N$ curve parameters $\l q$ and where: %  at the previous step
%One remarks that this system is well defined as soon as each parameter $\l p$ has an independent (and non null) action on the curve $\mathcal C$ shape. 
%
\bq
	2\psi(\l p)
	&=& 
	\int_{-1}^1\int_0^1\left(F(\vec X(x_1,x_2,\l p)) - g(x_2)\right)^2 dx_1 dx_2\label{eq:psi}\\
	\frac{\partial \psi}{\partial \l p} 
	&=& 
	\int_{-1}^1\int_0^1
	\left( \frac{\partial F}{\partial \vec X} \cdot \frac{\partial \vec X}{\partial \l p} \right)
	(f- g)
	dx_1 dx_2
	\label{eq:dpsidlp}\\
	\frac{\partial^2 \psi}{\partial \l p\partial \l q} 
	&=&
	\int_{-1}^1\int_0^1
	\left[
	\left(
	\frac{\partial \vec X}{\partial \l p}\cdot \frac{\partial^2 F}{\partial \vec X^2} \cdot \frac{\partial \vec X}{\partial \l q}
	+
	\frac{\partial F}{\partial \vec X} \cdot \frac{\partial^2 \vec X}{\partial \l p\partial \l q}
	\right)
	(f- g) \right. + \nonumber\\%dx_1 dx_2
	&& % \int_{-1}^1\int_0^1
	\quad \quad \left. 
	\left( \frac{\partial F}{\partial \vec X} \cdot \frac{\partial \vec X}{\partial \l p} \right)
	\left( \frac{\partial F}{\partial \vec X} \cdot \frac{\partial \vec X}{\partial \l q} \right)
	\right]
	dx_1 dx_2
	\label{eq:d2psidl2}
\eq
Annex \ref{sec:annex:magnitude} shows that it is possible, under some reasonable assumptions, to take into account only the last term (as done in DIC \cite{Hild2006}) thus:
\bq
	\frac{\partial^2 \psi}{\partial \l p\partial \l q} &\simeq& \int_{-1}^1\int_0^1 
	\left( \frac{\partial F}{\partial \vec X} \cdot \frac{\partial \vec X}{\partial \l p} \right)
	\left( \frac{\partial F}{\partial \vec X} \cdot \frac{\partial \vec X}{\partial \l q} \right)
	dx_1 dx_2\label{eq:d2psidl2s}
\eq
in which, from Eq.~(\ref{eq:revmap}):
\bq
%	\frac{\partial \vec X}{\partial x_1} &=& \left\| \frac{\partial \vec X^c}{\partial x_1} \right\| (1+\rho R x_2) \vec{e}_s \label{eq:dXdx1}\\
%	\frac{\partial \vec X}{\partial x_2} &=& R \vec{e}_r \label{eq:dXdx2}\\
	\frac{\partial \vec X}{\partial \l p} &=& \frac{\partial \vec X^c}{\partial \l p} 
	-
	R x_2 
	\left\| \frac{\partial \vec X^c}{\partial x_1} \right\|^{-1}
	\left(\frac{\partial^2\vec X^c}{\partial \l p\partial x_1}\cdot\vec e_r\right)
	\vec e_s
	\label{eq:dXdlp}
%	\\
%	\frac{\partial^2 \vec X}{\partial \l p\partial \l q} &=&
%	\frac{\partial^2 \vec X^c}{\partial \l p\partial \l q}
%	+ Rx_2 \frac{\partial^2\vec e_r}{\partial \l p\partial \l q}\\
%	\frac{\partial^2\vec e_r}{\partial \l p\partial \l q} &=&
%	-\left\| \frac{\partial \vec X^c}{\partial x_1} \right\|^{-1} \left( \frac{\partial^3 \vec X^c}{\partial \l p\partial \l q\partial x_1}\cdot \vec e_r \right) \vec e_s\nonumber\\
%	&& +\left\| \frac{\partial \vec X^c}{\partial x_1} \right\|^{-2}
%	\left[
%	\left(\frac{\partial^2\vec X^c}{\partial \l p\partial x_1}\cdot\vec e_r\right)
%	\left(\frac{\partial^2\vec X^c}{\partial \l q\partial x_1}\cdot\vec e_s\right)
%	\vec e_s\right.\nonumber\\
%	&&+\left(\frac{\partial^2\vec X^c}{\partial \l p\partial x_1}\cdot\vec e_s\right)
%	\left(\frac{\partial^2\vec X^c}{\partial \l q\partial x_1}\cdot\vec e_r\right)
%	\vec e_s\nonumber\\
%	&&-\left.\left(\frac{\partial^2\vec X^c}{\partial \l p\partial x_1}\cdot\vec e_r\right)
%	\left(\frac{\partial^2\vec X^c}{\partial \l q\partial x_1}\cdot\vec e_r\right)
%	\vec e_r\right]\label{eq:d2erdlpdlq}
\eq
The derivatives of the curve points $\vec X^c$ are supposed either analytically or numerically known.
Curvilinear abscissa $s$ and curvature $\rho$ are:
\bq
	s &=& \int_0^{x_1} \left\|\frac{\partial \vec X^c}{\partial \xi_1}\right\| d \xi_1 \label{def:s}\\
	\rho &=& -\left(
	\frac{\partial^2 \vec X^c}{\partial x_1^2} \cdot \vec{e}_r\right)
	\left\|\frac{\partial \vec X^c}{\partial x_1}\right\|^{-2}\label{def:rho}
\eq
and $L=s(1)$ is the overall curve length.
If the non-overlapping condition:
\be
	|\rho| R < 1 \label{cond:nonoverlap}
\ee
is not fulfilled, the center of the osculating circle of $\mathcal C$ of radius $1/|\rho|$ is inside the virtual image G, thus some points in the vicinity of this center are defined at least twice.
% (for different couples $(x_1,x_2)$).
However, in a practical point of view, experience shows that it is possible to overcome this second condition as soon as the sharp corners of $\mathcal C$ do not exceed the right angle because sharper angles put in coincidence inner black points of G with outer white points of F.
At last, from Eq.~(\ref{eq:dXdlp}) the curve must not have any stationary points:
\be
	\left\| \frac{\partial \vec X^c}{\partial x_1} \right\| > 0\label{eq:nonstat}
\ee
%
% Conditions in Eqs.~(\ref{cond:nonoverlap}, \ref{eq:nonstat}) correspond to a locally bijective mapping.
% but the full bijection requires the curve do not passes at a distance less than $2R$ of itself.
% thus one should avoid silhouettes with crossing or isthmus thinner than $2R$.

%---------------------------------------------------------------
\section{Uncertainty of the VIC measurement}\label{sec:2}
%---------------------------------------------------------------

%---------------------------------------------------------------
\subsection{Set of simplified equations in ideal cases}\label{sec:idcases}
In order to study the precision of the method, the above set of equation is simplified hereafter.
At first we suppose that, close to the solution, $F(\mathbf{X}) \simeq f(x_2)$ (as it is the case for $g$, see Eqs.~(\ref{eq:Gg}, \ref{eq:gdexsil})) thus:
\bq
%	F(\mathbf{X}) &=& f(x_2)\label{hyp:fx2}\\
	\frac{\partial F}{\partial \vec X} &=& \frac{f'}{R}\vec{e}_r\label{eq:dfdX}
\eq
Together with Eq.~(\ref{eq:dXdlp}), this allows the separation of variables in Eq.~(\ref{eq:dpsidlp}):
\be
	\frac{\partial \psi}{\partial \l p} 
	= 
	\frac{1}{R}
	\int_0^1
	\frac{\partial \vec X^c}{\partial \l p}
	\cdot
	\vec{e}_r\,
	dx_1
	\int_{-1}^1
	f'(f- g)
	dx_2
	\label{eq:dpsidlp_fx2}
\ee
The current term of the first integral is null if ${\partial \vec X^c}/{\partial \l p}$ is everywhere collinear to $\vec e_s$, corresponding to a tangential motion which lets the curve unchanged: such case has to be avoided when choosing a curve equation. 
%(for example this would have been the case if one had introduced a angle parameter in the circle's equation (/ref{circle})). 
Thus, at convergence of the Newton scheme, when ${\partial \psi}/{\partial \l p}=0$:
\be
	\int_{-1}^1
	f'(f- g)
	dx_2=0
	\label{eq:conv}
\ee
%
%The virtual image gray level is such as $g-1/2$ is an odd function. 
%This means that the VIC identifies the axis of the function $f-1/2$ if this one is odd (because in this case the integral over $x_2$ is automatically null), whatever the arbitrary parameter $R$ (the half-width of the virtual image) which defines, in the present coordinates $x_2$, the width of the support of the function f.
At last, if the virtual image borders lie one in the white background and one in the black silhouette: $f(-1)=0$ and $f(1)=1$, an integration by parts gives:
%
%\be
%	f(-1)=0\quad\mathrm{and}\quad f(1)=1\label{hyp:cont}
%\ee
%and integral in Eq.~(\ref{eq:conv}) can be expressed thanks to Eq.~(\ref{eq:gdexsil}) and an integration by parts as: %  ($f-1/2$ being not necessary odd)
%
\be
	\frac{1}{2}
	\int_{-1}^1 \left( f(x_2)-\frac{1}{2} \right) dx_2 = 0\label{eq:vicint}
\ee
which shows that, at convergence, the mean value of $f$ is $1/2$.
\subsection{Exact 1D discrete measurement}\label{sec:effectdiscret}

% We show hereafter that the VIC, in one dimension, recovers the exact location of an edge located at any point $X_0$ between two pixels (Fig.~\ref{fig:1D_vision}).
%
\begin{figure}[h]
\begin{center}
\includegraphics[scale=0.50]{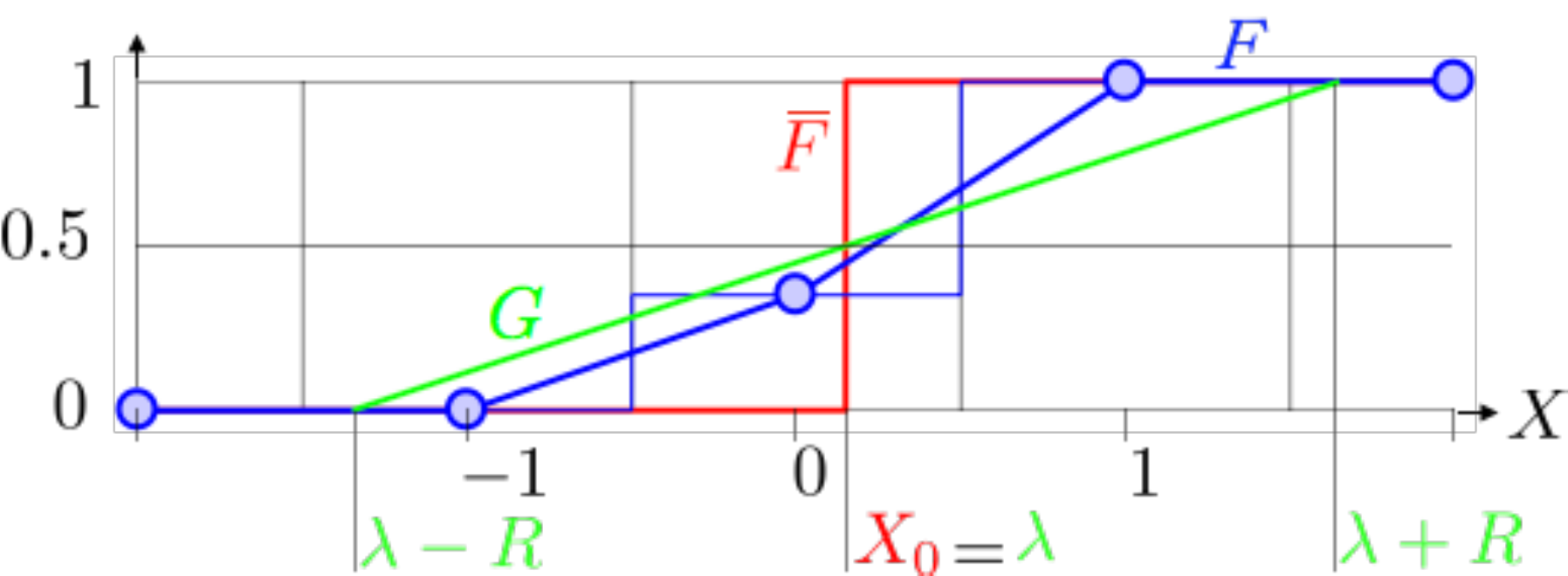}
\caption{1D VIC. Luminance $\bar F$ (red), digital image and its linear interpolation $F$ (blue), virtual image $G$ at best correlation (green)}
% for $R=0,5$ pixel and $X_0=2,1$ pixel
\label{fig:1D_vision}
\end{center}
\end{figure}
Let $\bar F(X)=H(X-X_0)$, where $H$ is the Heavyside distribution, be the physical luminance of a 1D silhouette (Fig.~\ref{fig:1D_vision}).
Supposing ideal sensor (linear and homogenous) and optics, the $i^\mathrm{th}$ pixel returns the value $\int_{i-1/2}^{i+1/2} \bar F dX$ (blue dots). 
%In particular, if the pixel that contains $X_0$ corresponds to $i=0$, the physical 1D image (blue circles on Fig~\ref{fig:1D_vision}) is:
%%
%\bq
%	F(i<0) &=& 0\nonumber\\
%	F(0) &=& \frac{1}{2}-X_0\nonumber\\
%	F(i>0) &=& 1
%\eq
%%
%The linear interpolation of the previous function (blue segments on Fig~\ref{fig:1D_vision}) is:
%%
%\bq
%	X<-1 &\Longrightarrow& F(X) = 0\nonumber\\
%	X\in[-1,0[&\Longrightarrow& F(X) = \left(\frac{1}{2}-X_0\right)(X+1)\nonumber\\
%	X\in[0,1[&\Longrightarrow& F(X) = \left(\frac{1}{2}+X_0\right)(X-1)+1\nonumber\\
%	X\geqslant 1 &\Longrightarrow& F(X) = 1\label{eq:interpol}
%\eq
In 1D, the curve $\mathcal C$ is degenerated into a point ($x_1$ is meaningless) whose parameterized equation is simply chosen as $X^c = \lambda$ and the mapping (Eq.~\ref{eq:revmap}) reduces to $X=\lambda+Rx_2$.
Due to the absence of curvature, expressions $\Psi$ (Eq.~\ref{eq:grandpsi}) and $\psi$ (Eq.~\ref{eq:petitpsi}) are equivalent thus Eq.~(\ref{eq:vicint}) corresponds to:
\be
	\frac{1}{2R}
	\int_{\lambda-R}^{\lambda+R}\left( F(X)-\frac{1}{2} \right) dX = 0,\label{eq:convinX}
\ee
if $f(-1)=F(\lambda-R)=0$ and $f(1)=F(\lambda+R)=1$, \emph{i.e.} if the support of $G$ is wide enough: $\lambda-R<-1$ and $\lambda+R>1$.
Because $-0.5<\lambda<0.5$, this leads to impose $R>1.5$.
Solving this integral with the analytical expression $F(X)$ of the linear interpolation (thick blue segments in Fig.~\ref{fig:1D_vision}) gives straightforwardly $\lambda=X_0$.
This shows that the VIC measurement $X^c$ corresponds exactly to the prescribed edge location $X_0$, whatever $X_0$ and $R>1.5$.
%, $\lambda$ corresponds exactly to the researched value $X_0$, whatever $X_0$ and $R>1.5$. 
%This 1D reasoning can be extended for the 2D curves if we suppose that f depends only upon $x_2$ thus that the separated form in Eq.~(\ref{eq:dpsidlp_fx2}) is still available. 

%---------------------------------------------------------------
\subsection{Uncertainty due to curve mismatch and local correlation indicator}\label{sec:localindic}

\begin{figure}[h]
\begin{center}
\includegraphics[scale=0.55]{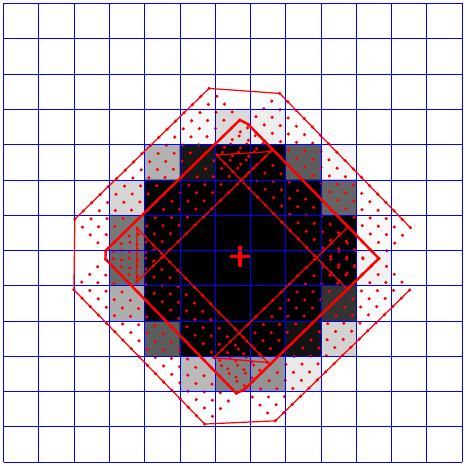}$\quad$\includegraphics[scale=0.75]{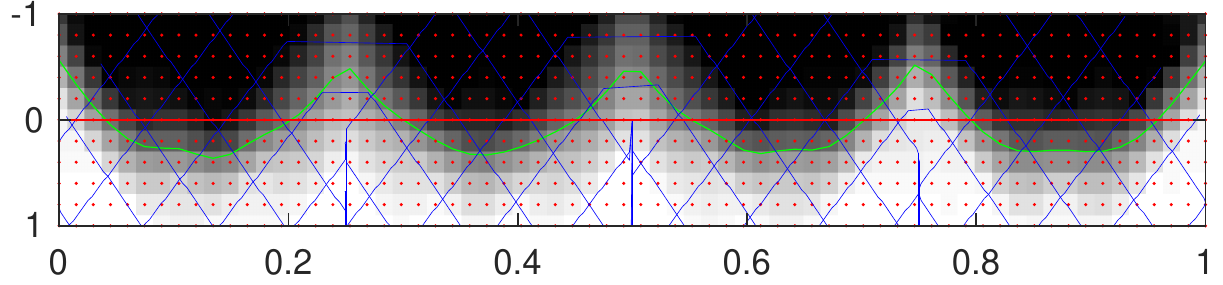}\\
\vspace{0.2in}
\includegraphics[scale=0.55]{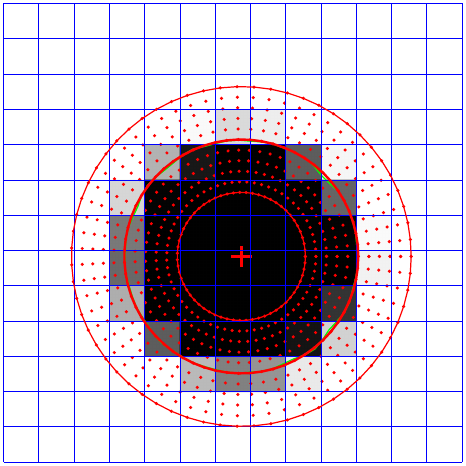}$\quad$\includegraphics[scale=0.75]{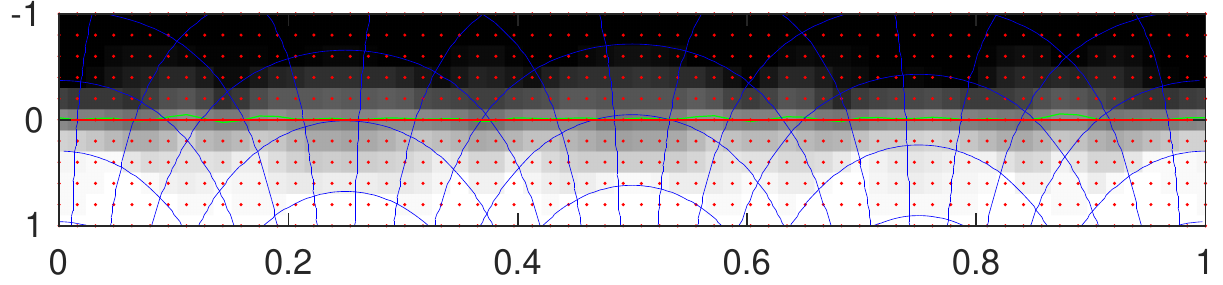}
\caption{Images F (left) and f (right) of a small disc identified by a square (top) and a circle (bottom). 
Pixel edges (blue), computational points (red) and correlation indicator $\mu$ (green)}
\label{fig:fff}
\end{center}
\end{figure}

The VIC method requires the user-chosen curve $\mathcal C$ to be able to fit the contour of interest.
Fig.~\ref{fig:fff} shows that, if the curve matches, f appears as invariant along $x_1$ (very similar to g) but shows waviness in the opposite case.
However, a more objective indicator is necessary to quantify the quality of the identification.
A straightforward idea consists in using $\partial \psi/\partial x_1$ as local correlation function
%\be
%	\varphi(x_1) = \frac{1}{2}\int_{-1}^1 (f(x_1,x_2)-g(x_2))^2\,dx_2,
%\ee
but, g being not physical and $R$ being user chosen, this function only brings a qualitative information and does dot distinguish if the curve is inside or outside the contour.

From Eq.~(\ref{eq:vicint}), we define the signed distance:
%
%Eq.~(\ref{eq:vicint}) shows that the average of $f$ along $x_2$ is $1/2$ in 1D cases. % <f(x_2)>=
%This is also true in 2D cases, for all $x_1$, if the curve is everywhere able to match the contour.
%The signed distance 
%
\be
	\mu(x_1) = \int_{-1}^1 \left( \frac{1}{2}-f(x_1,x_2) \right) dx_2\label{eq:defmu}
\ee
This one is defined in the frame $\vec x$ and corresponds to $\mu R$ in the pixel frame $\vec X$.
Fig.~\ref{fig:fff} shows that $\mu(x_1)$ represents a local identification of the boundary.\\

\begin{figure}[h!]
\begin{center}
\includegraphics[scale=0.75]{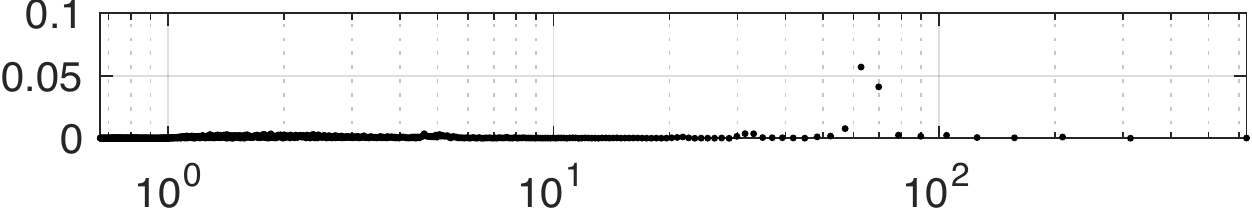}a)\\
\includegraphics[scale=0.75]{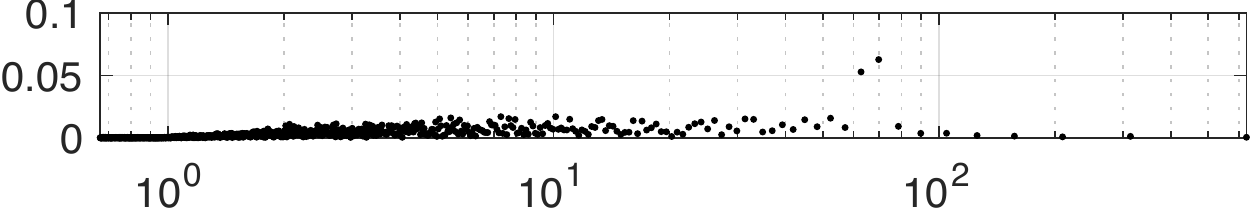}b)\\
\includegraphics[scale=0.75]{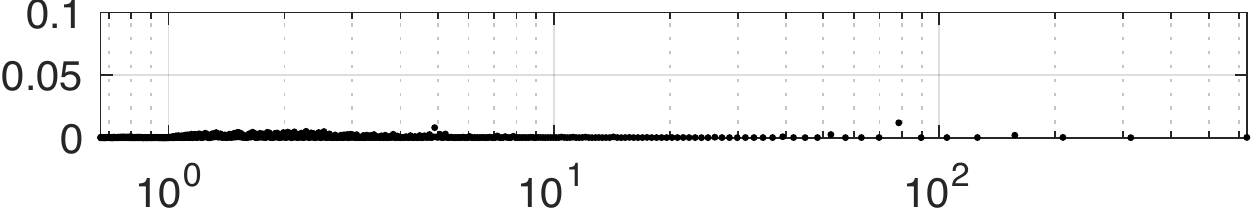}c)\\
\includegraphics[scale=0.75]{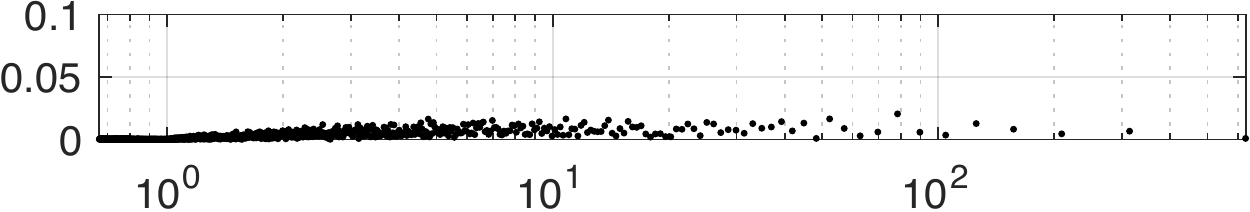}d)
\caption{FFT of $R\mu$ for a 100 pixel radius disc with (a) $\sigma_0=0$ and 10 points B-Spline, (b) $\sigma_0=0.1$ and 10 points B-Spline, (c) $\sigma_0=0$ and circle,  (d) $\sigma_0=0.1$ and circle.}
\label{fig:FFTdelta}
\end{center}
\end{figure}
Fig.~\ref{fig:FFTdelta} is related to the well-know impossibility for a B-Spline to depict a circle \cite{Piegel-Tiller89}.
When using a 10 points B-Spline, a peak  of magnitude $\simeq 0.05$ pixel at wavelength $63$ pixels ($\simeq L/10$ pixel) is visible in the FFT of $R\mu$.
It reveals a periodic oscillation of the curve from inside to outside of the exact circle in between the ten (regularly spaced) control points, which would be hard to see on a representation such has Fig.~\ref{fig:fff}.
Cases with noisy images lead to an additional noise spectrum whose mean amplitude is close to the estimation of Eq.~(\ref{eq:sigma}) of $21\times10^{-3}$ for the B-Spline and $8\times10^{-3}$ for the circle.\\

Curve mismatching induces long wave oscillations of $R\mu$ which are revealed by a spectral analysis as long as they are not hidden by image noise.
The acceptable precision, \emph{i.e.} the magnitude of the maximum peak, remains the the user's decision. 
%However, values greater than $R$ indicate that the virtual image support does not contain the contour at some points.

%---------------------------------------------------------------
\subsection{Uncertainty associated to discretization}\label{sec:numerics} % Merged

If the VIC has been shown in Sec.~\ref{sec:effectdiscret} to be theoretically exact in 1D, things are more complicated in 2D.
The pixel grid is used as computational frame in most of image analysis methods, including DIC and some versions of VIC using $\Psi$ \cite{francois13exm}. 
However, many tests showed that computing on a regular discretization of $(x_1,x_2)$ (see Fig.~\ref{fig:fff}) provides better precision, as soon as the distance between two corresponding points $(X_1,X_2)$ is less than $1/3$ pixel \cite{francois11epj}.\\

The values of F, required for Eq.~(\ref{eq:petitpsi}) at these non integer values are obtained by interpolation.
Another series of tests showed that the simplest and fastest linear interpolation gives equivalent or even better precision than cubic or B-spline interpolation.
This is different from DIC, but in accordance with the analytical analysis in Sec.~\ref{sec:effectdiscret}.\\

\begin{figure}[h!]
\begin{center}
\includegraphics[scale=0.35]{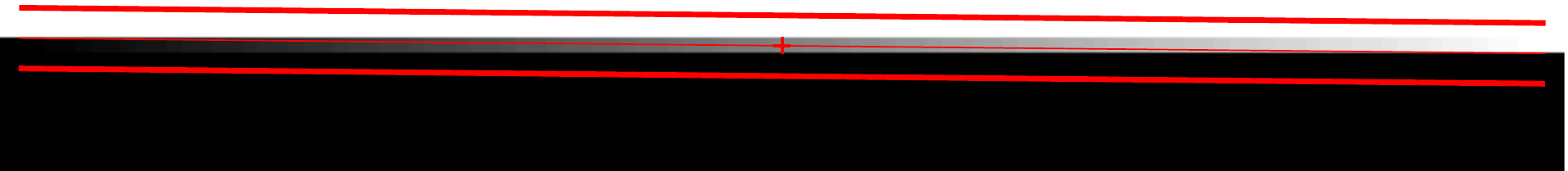}
\caption{Line segment test with small angle of inclination}% at small angle
\label{fig:bench}
\end{center}
\end{figure}
In Sec.~\ref{sec:effectdiscret} we showed that $R$ should be greater than 1.5 pixel. In order to set the optimal value of $R$ for 2D images, we proceed to tests on synthetic images of linear edges identified with a line segment (Fig.~\ref{fig:bench}).
To address any cases, the angle $\theta$ from $X_1$ to the edge is varied from 0 to $\pi/4$, the ordinate of its midpoint is varied of $\Delta X_{02} \in [0, 1[$ pixel and the length $L=100$ of the segment is varied of $\Delta L\in[0, 1[$ pixel.
Ten levels are used for each variation.
The mean $m_d$ and the standard deviation $\sigma_d$ of the distance $\delta(x_1)$ between identified and exact segments are computed.
In average, setting $R=1.5$ gives $m_d= -4.09\times 10^{-5}$ and $\sigma_d=3.76\times 10^{-4}$ and setting $R=2$ gives $m_d=-5.99\times 10^{-5}$, $\sigma_d=7.92\times 10^{-4}$.\\

\begin{figure}[h!]
\begin{center}
\includegraphics[scale=0.175]{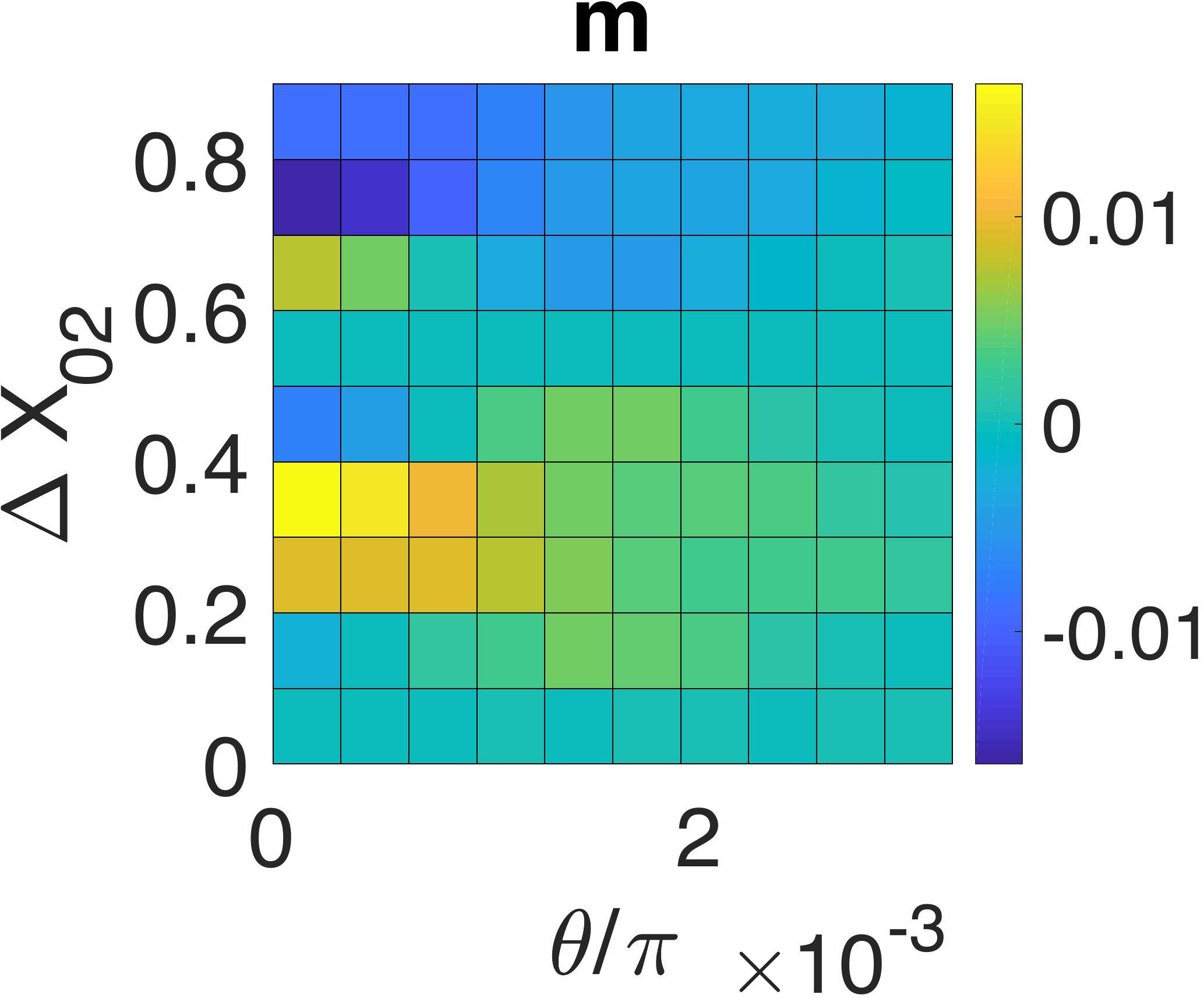}\hspace{0,25in}\includegraphics[scale=0.175]{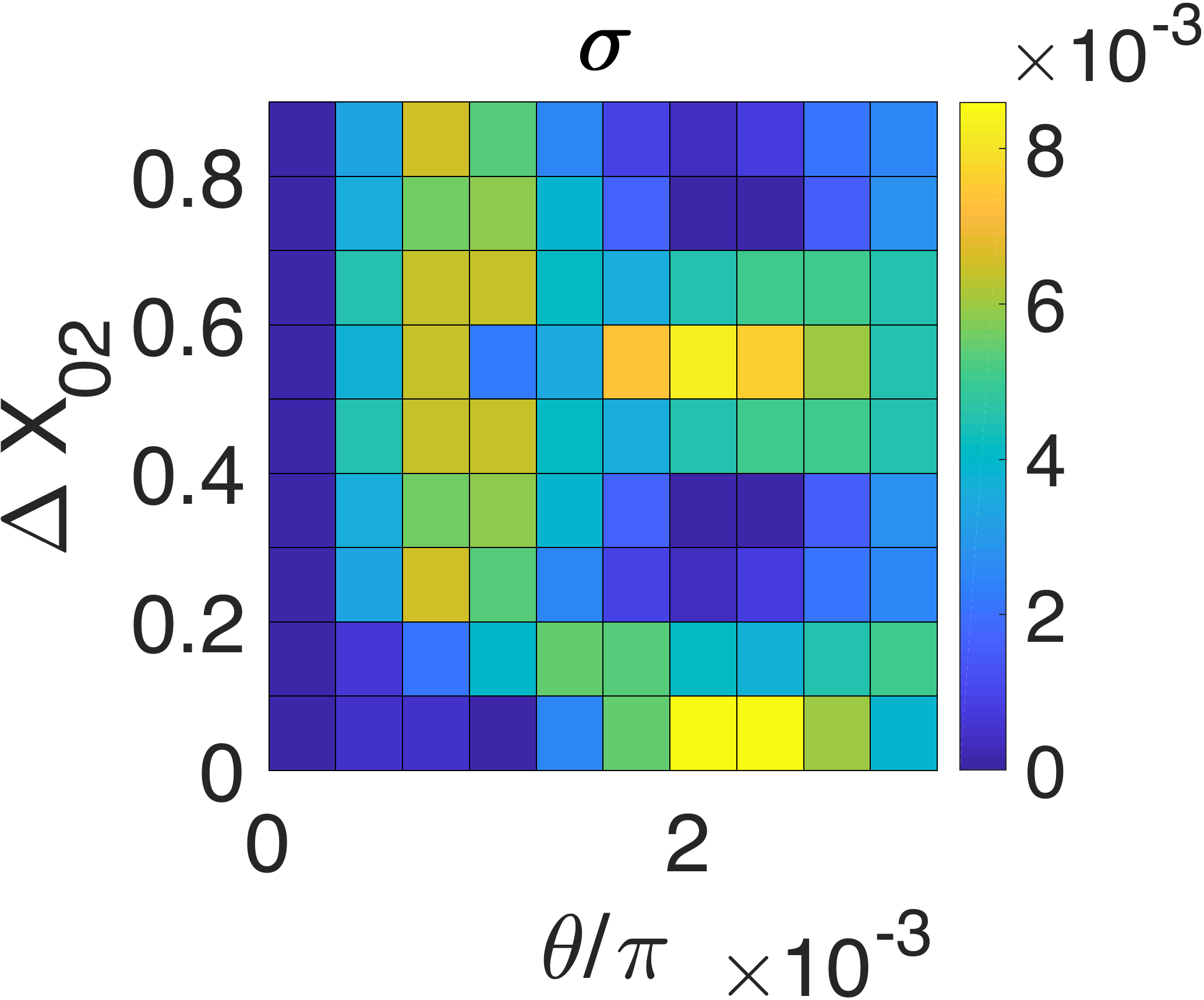}\\
\includegraphics[scale=0.175]{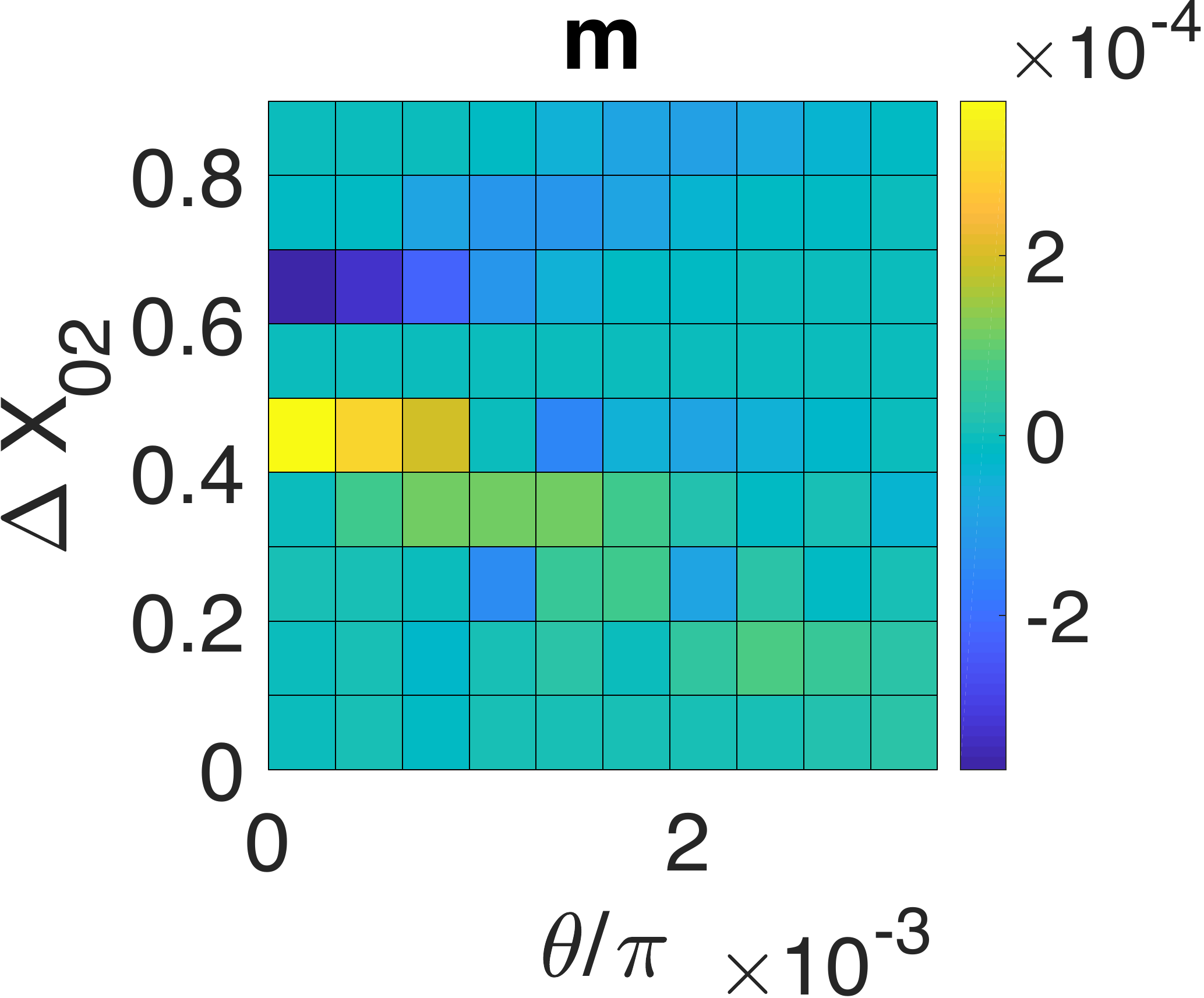}\hspace{0,25in}\includegraphics[scale=0.175]{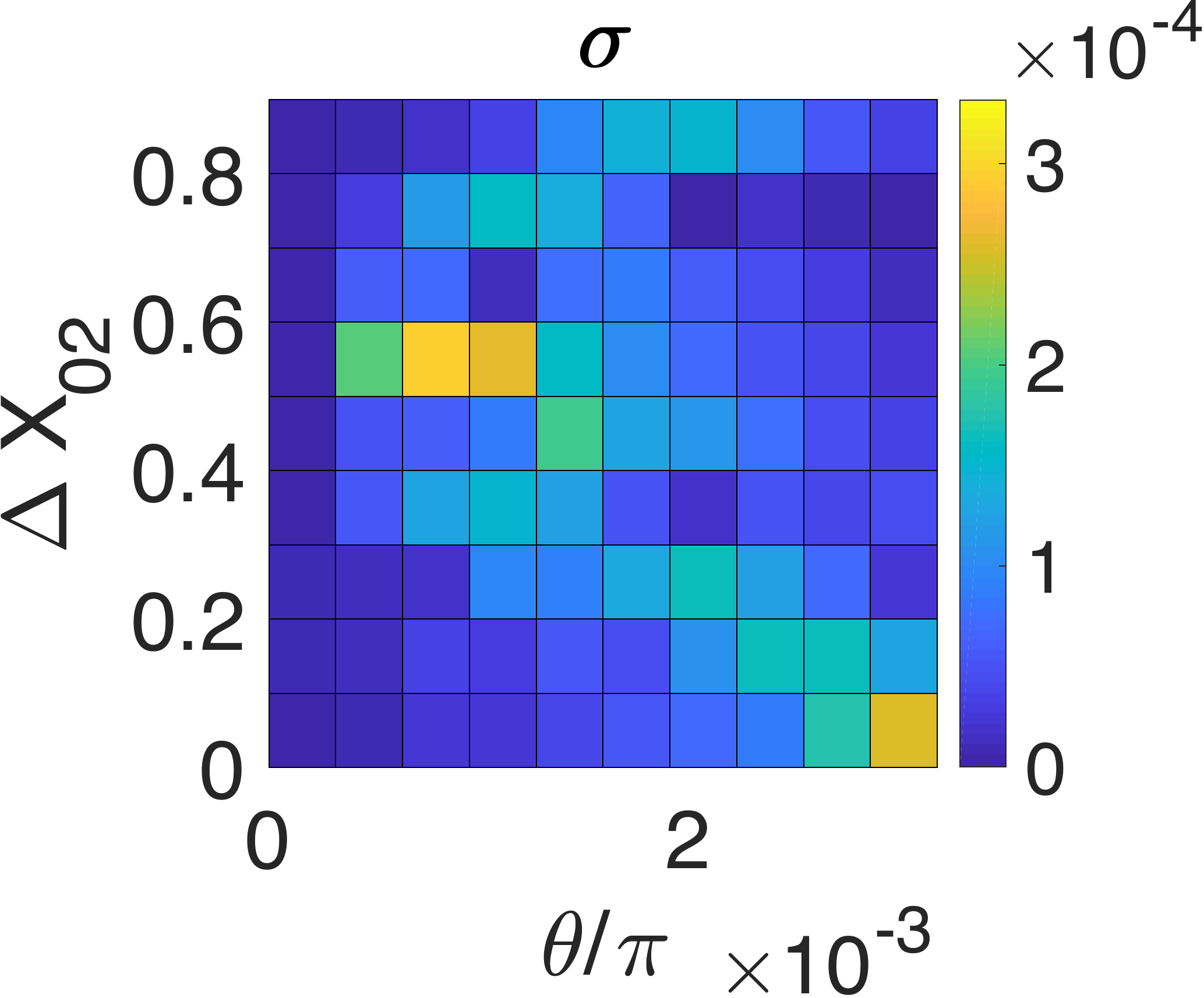}\\
\caption{Statistics on the line segment test at small angles for $R=1.5$ (top) and $R=2$ (bottom)}
\label{fig:moyDL}
\end{center}
\end{figure}

However Fig.~\ref{fig:moyDL} shows that choosing $R=2$ eliminates pathologic cases of small angles (such as shown by Fig.~\ref{fig:bench}).
Setting other (especially larger) values for $R$ did not provide any advantage.
As a consequence, $R=2$ appears to be the optimal value for the VIC.\\

In order to get an estimator of the uncertainty associated to the sole effect of discretization, similar line segment tests have been realized, varying the length $L$ over two decades.%$L\in\{5,10,20,40,80,160,320,640,1280\}$ pixels.
\begin{figure}[h!]
\begin{center}
\includegraphics[scale=0.30]{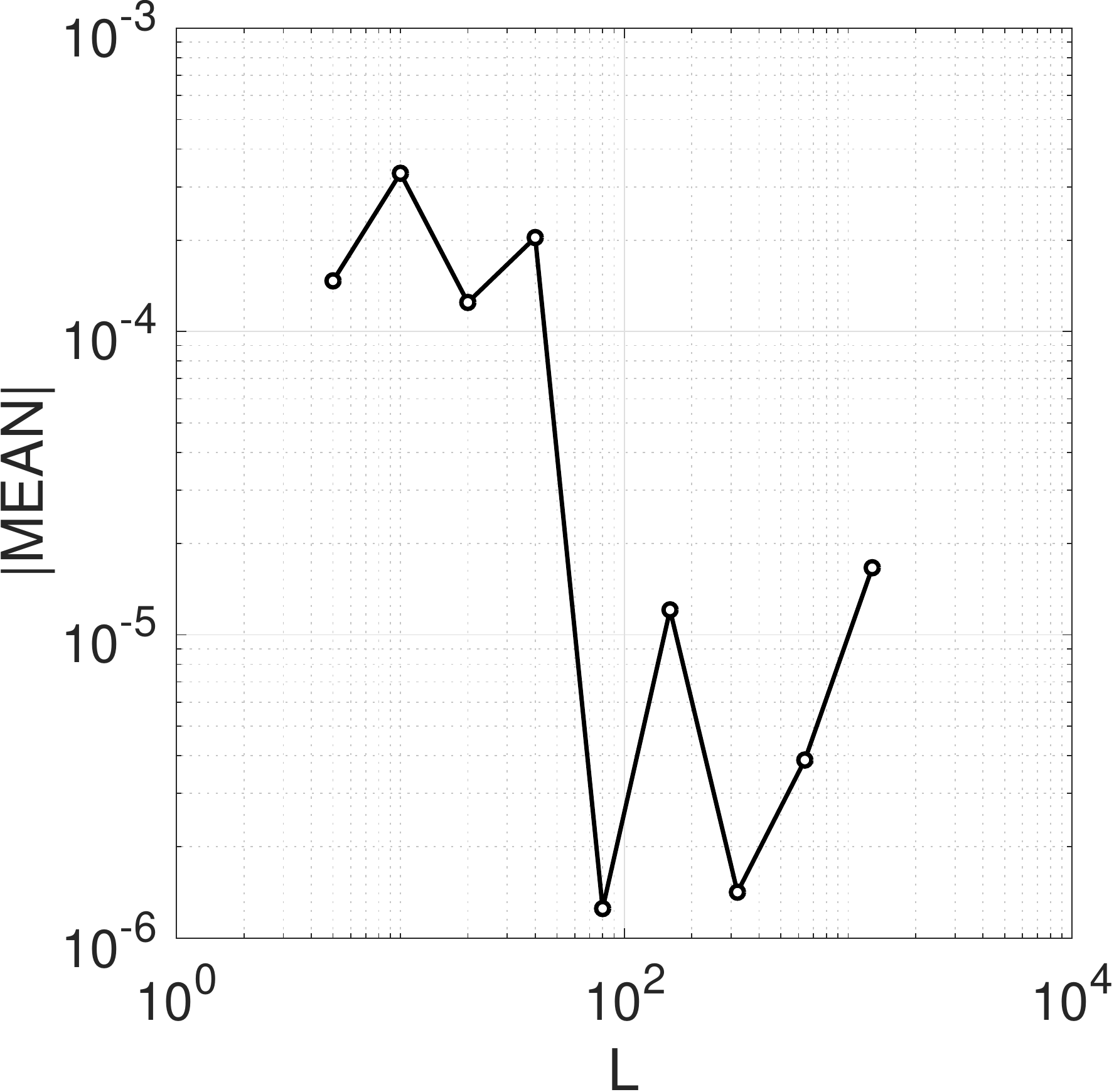}\includegraphics[scale=0.30]{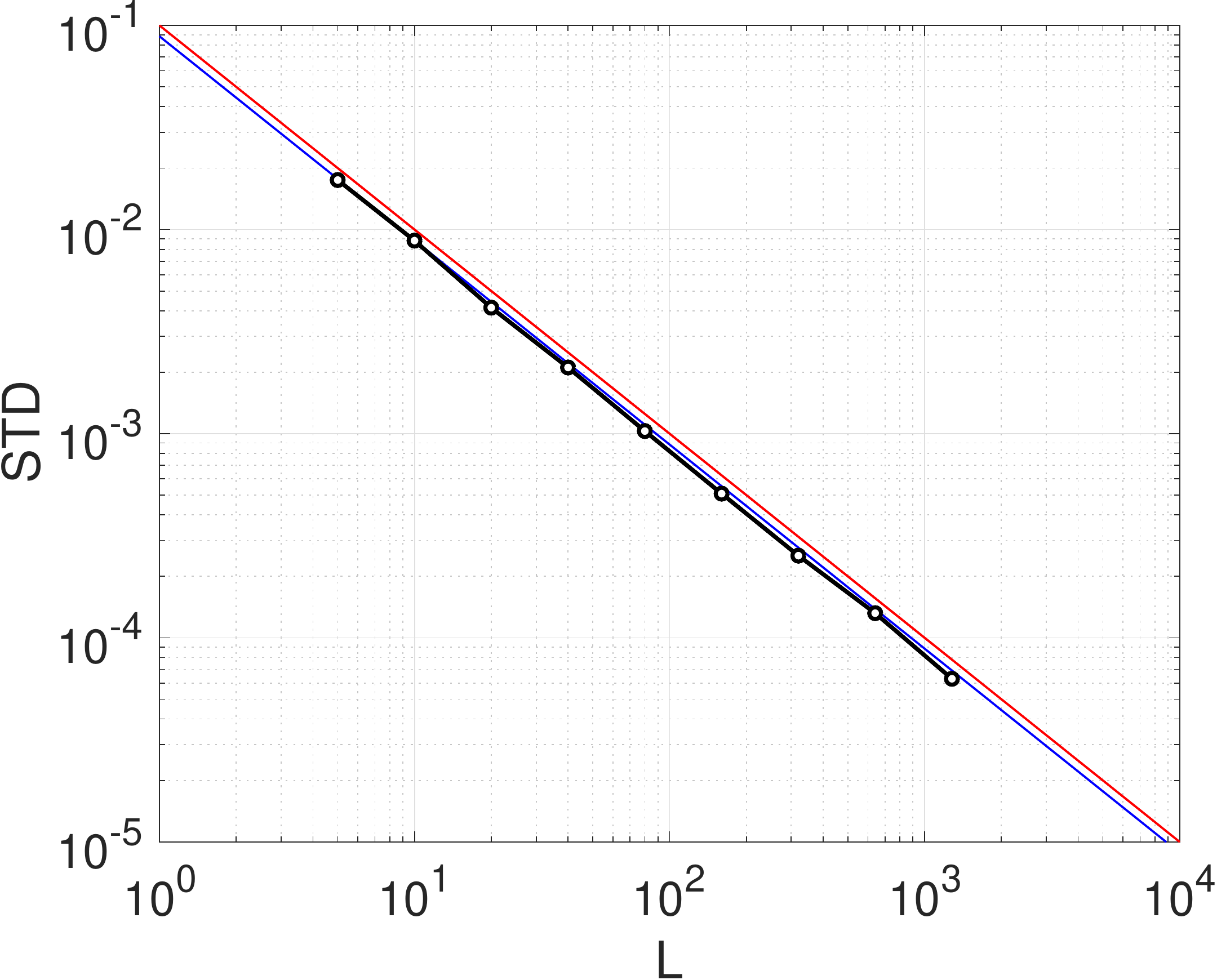}
\caption{Statistics of the line segment test with variable length}
\label{fig:segstat}
\end{center}
\end{figure}
Fig.~\ref{fig:segstat} shows that the mean distance $m_d$ is weak for all $L$ and that the standard deviation $\sigma_d$ is very close to its linear regression $\sigma_d = {0.0885}/{L}$ (blue line on Fig.~\ref{fig:segstat}).
The proposed empirical rule:
\be
	\sigma_d\simeq \frac{N}{20 L}\label{eq:sigma_disc}
\ee
corresponds to the red line.
The role of the number of curve parameters $N$ in this equation, of two in these tests (ordinate and angle), is supposed to be equivalent to the one it has in Eq.~(\ref{eq:sigma}).
%This fit corresponds to the red line in Fig.~\ref{fig:segstat}.
%The role of $N$, the numbers of degree of freedom of the curve, is similar in this equation to its role in Eq.~(\ref{eq:sigma}), for the case of image noise.% in next Sec~\ref{sec:noise} 

%---------------------------------------------------------------
\subsection{Uncertainty associated to brightness and contrast defects}\label{sec:brightness}

\begin{figure}[htbp]
	\begin{center}
		\includegraphics[scale=0.60]{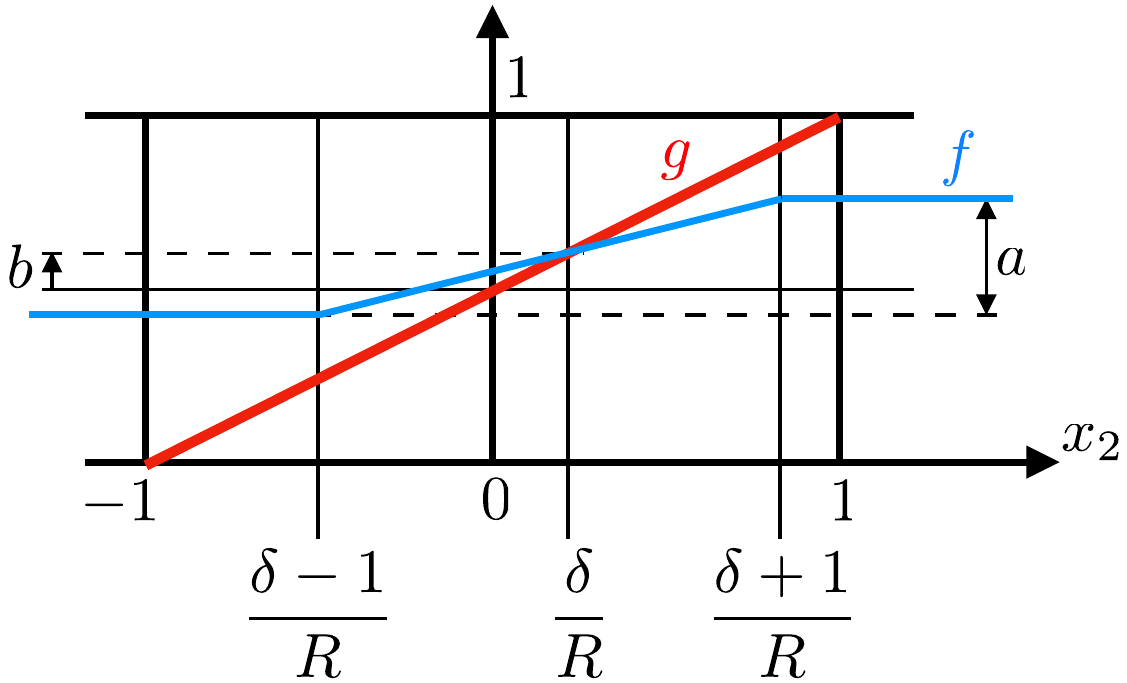}
		\caption{Study of amplitude $a$ and bias $b$ effects}
		\label{fig:amplitude_bias}
	\end{center}
\end{figure}

At ideal, the silhouette is black and the background is white. However, real images may have contrast and luminance deviations. 
From Sec.~\ref{sec:effectdiscret}, the average of $F(X)$ over all possible $X_0\in[-0.5,0.5]$ is linear: $<F>=({X+1})/{2}$.
Thus, for this analytical study, we suppose $f$ to be the continuous linear piecewise function (see Fig.~\ref{fig:amplitude_bias}):
\bq
	x_2<\frac{\delta-1}{R} & \Longrightarrow & f(x_2)=b+\frac{1-a}{2}\nonumber\\
	\frac{\delta-1}{R}\leq x_2<\frac{\delta+1}{R} & \Longrightarrow & f(x_2)=a\frac{Rx_2-\delta}{2}+b+\frac{1}{2}\nonumber\\
	\frac{\delta+1}{R}\leq x_2 & \Longrightarrow & f(x_2)=b+\frac{1+a}{2}
	\label{eq:linf}
\eq
where $a$ is the amplitude (contrast), $b$ the bias (luminance), $\delta/R$ the location of the researched edge.
The origin of the virtual image at $x_2=0$ defines the VIC measurement thus $\delta/R$ (in the frame $\vec x$) or $\delta$ (in the pixel frame $\vec X$) is the measurement error. 
%From Eq.~(\ref{eq:revmap}) degenerated in 1D, $\delta$ (pixel) is the VIC measurement error.
%
Used together with Eq.~(\ref{eq:conv}), which is fulfilled at convergence, above expressions give:
\bq
	\delta = 2Rb\label{eq:petitdelta}
\eq
As a consequence, the contrast $a$ has no influence on the precision but a luminance variation $b$ induces a bias $\delta$ in the measurement.
% This shows that the amplitude $a$ has no influence on the precision of the measurement but that the bias $b$ induces a deviation, proportional to $b$ and of maximum amplitude $R$ in the limit case $b=1/2$.
In a practical point of view, this is easily annulated during computation by a linear correction of the gray levels of F.
% set in order to obtain $b=0$ (and possibly $a=1$).
%
%\bq
%	\frac{F}{f_w- f_k} - \frac{f_k}{f_w - f_k} &\to& F %\\
%%	f_w &=& <f(x_1,1)>\\
%%	f_k &=& <f(x_1,-1)>
%\eq
%
%whose effect is to bring $f_w$, the average of $f(x_1,1)$ located in the background, to 1 and to bring $f_k$, the average of $f(x_1,-1)$ located in the silhouette, to 0 at the ideal.
Similar calculus shows that non linear image corrections should be avoided because they induce a bias.
In particular, best results will be obtained with CCD sensors with good linearity.
%In particular, better results are obtained by using linear CCD sensors.

%---------------------------------------------------------------
\subsection{Uncertainty associated to image noise}\label{sec:noise}

Inevitable image noise leads to uncertainty in the VIC measurement. 
We suppose now that each pixel of integer coordinates $(i,j)$ is the sum of an exact value $F_{ij}$ and a gaussian noise $N_{ij}$, spatially uncorrelated, of zero mean and of standard deviation $\sigma(N_{ij})=\sigma_0$. 
With the hypothesis of Sec.~\ref{sec:idcases}, we show in Annex \ref{sec:annex:noise} that:
\be
	\sigma\left(\frac{\partial \psi}{\partial \l p}\right)
	\simeq 
	\frac{\sigma_0}{R\sqrt{2RL}} \sqrt {\int_0^1 \left( \frac{\partial \vec X^c}{\partial \l p} \cdot \vec{e}_r \right) ^2\,d x_1}
	\label{eq:sigmadpsidlp}
\ee
From the Newton scheme (Eq.~\ref{eq:newton}), this term is associated to the standard deviation of each shape parameter by:
\be
	\sigma^2\left(\frac{\partial \psi}{\partial \l p}\right) = 
	\left(\frac{\partial^2 \psi}{\partial \l p\partial \l q}\right)^2 \sigma^2(\l q)
	\label{eq:std}
\ee
The average $\sigma_n$ of the standard deviation of the distance $\vec X^c \cdot \vec{e}_r$ from the measurement to the noiseless solution is:
\be
	\sigma_n^2 = \int_0^1\sigma^2 \left(\vec X^c \cdot \vec{e}_r \right) \, d x_1\label{eq:defmu}
\ee
From $d \vec X^c  = ({\partial \vec X^c}/{\partial \l q})\, d \l q$ we deduce:
\be
	\sigma^2\left(\vec X^c \cdot \vec{e}_r \right) = \sum_q \left( \frac{\partial \vec X^c}{\partial \l q} \cdot \vec{e}_r \right)^2 \sigma^2(\l q)\label{eq:sigma2}
\ee
Gathering previous expressions gives $\sigma_n^2$. %
%\be
%	\sigma_n^2 = 
%	\frac{\sigma_0^2}{2LR^3}
%	\int_0^1 \left( \frac{\partial \vec X^c}{\partial \l p} \cdot \vec{e}_r \right)^2 d x_1
%	\left(\left[\frac{\partial^2 \psi}{\partial \l p\partial \l q}\right]^{-1}\right)^2
%	\int_0^1 \left( \frac{\partial \vec X^c}{\partial \l q} \cdot \vec{e}_r \right)^2 d x_1
%	\, d x_1
%\ee
%
However this complex expression has to be computed in each cases.
A simpler approximation is obtained by supposing at first a perfectly contrasted image with $a=1$ and $b=0$. Eq.~(\ref{eq:linf}) and Eq.~(\ref{eq:d2psidl2s}) give:%  (Section~\ref{sec:idcases})
\be
	\frac{\partial^2 \psi}{\partial \l p\partial \l q} = 
	\frac{1}{2R}\int_0^1 
	\left( \frac{\partial \vec X^c}{\partial \l p} \cdot \vec{e}_r \right)
	\left( \frac{\partial \vec X^c}{\partial \l q} \cdot \vec{e}_r \right)
	\,d x_1
\ee
At second we retain only the diagonal terms of this matrix, which corresponds to consider that each shape variable acts on a separate part of the curve. This gives a simple approximation of the VIC standard deviation due to the image noise:
%One finds:
%%
%\be
%	\sigma^2(\l q) = \frac{2\sigma_0^2}{RL
%	\int_0^1 \left( \frac{\partial \vec X^c}{\partial \l q} \cdot \vec{e}_r \right)^2 d x_1}
%\ee
%
%
\be
	\sigma_n = \sigma_0\sqrt{\frac{2N}{RL}}\label{eq:sigma}
\ee
The proportionality between $\sigma_n$ and the image noise $\sigma_0$ is common with DIC uncertainty analysis \cite{Roux2006,Sutton_09,Su2016,Bornert2018}. 
%Associated to an averaging effect, the larger $R$, the smaller $\sigma_n$, but Eq.~(\ref{eq:petitdelta}) shows that retaining a large $R$ increases the error $\sigma_n$ associated to the bias $b$. Furthermore it is shown in Sec.~\ref{sec:effectdiscret} that $R$ must be larger than $1.5$ pixels.
%Practically, due to local variation of luminance, the bias error is generally predominant, suggesting an optimal value $R=1.5$ pixel.
%For a given $R$, 
Doubling the image resolution doubles $L$ thus divides $\sigma_n$ by $\sqrt{2}$.
%The number of shape parameters $N$ is associated to the complexity, the flexibility, of the curve family.
The uncertainty $\sigma_n$ is proportional to $\sqrt{N}$: this weak dependance allows the user to retain complex curve families.
%The quality of the picture is not present in this expression because a perfectly neat and contrasted image of the edge was assumed.
% A good estimation of the actual noise level $\sigma_0$ can be obtained by computing the STD of the black and white borders $f(x_1,-1)$ and $f(x_1,1)$.
%Even for noiseless images, 
%  in case of uniform distribution and
This formula erroneously suggest the use of large $R$ but the calculus is valid for the active part of the virtual image thus one may consider $R\leq 2$. 
The quantification noise, of classical expression $\sigma_{0q}=(2^{nb}\sqrt{12})^{-1}$ where $nb$ is the bit depth, can also be taken into account as an additional image noise.

%---------------------------------------------------------------
\subsection{Summary}\label{sec:summary}

\begin{figure}[h!]
	\begin{center}
		\includegraphics[scale=0.325]{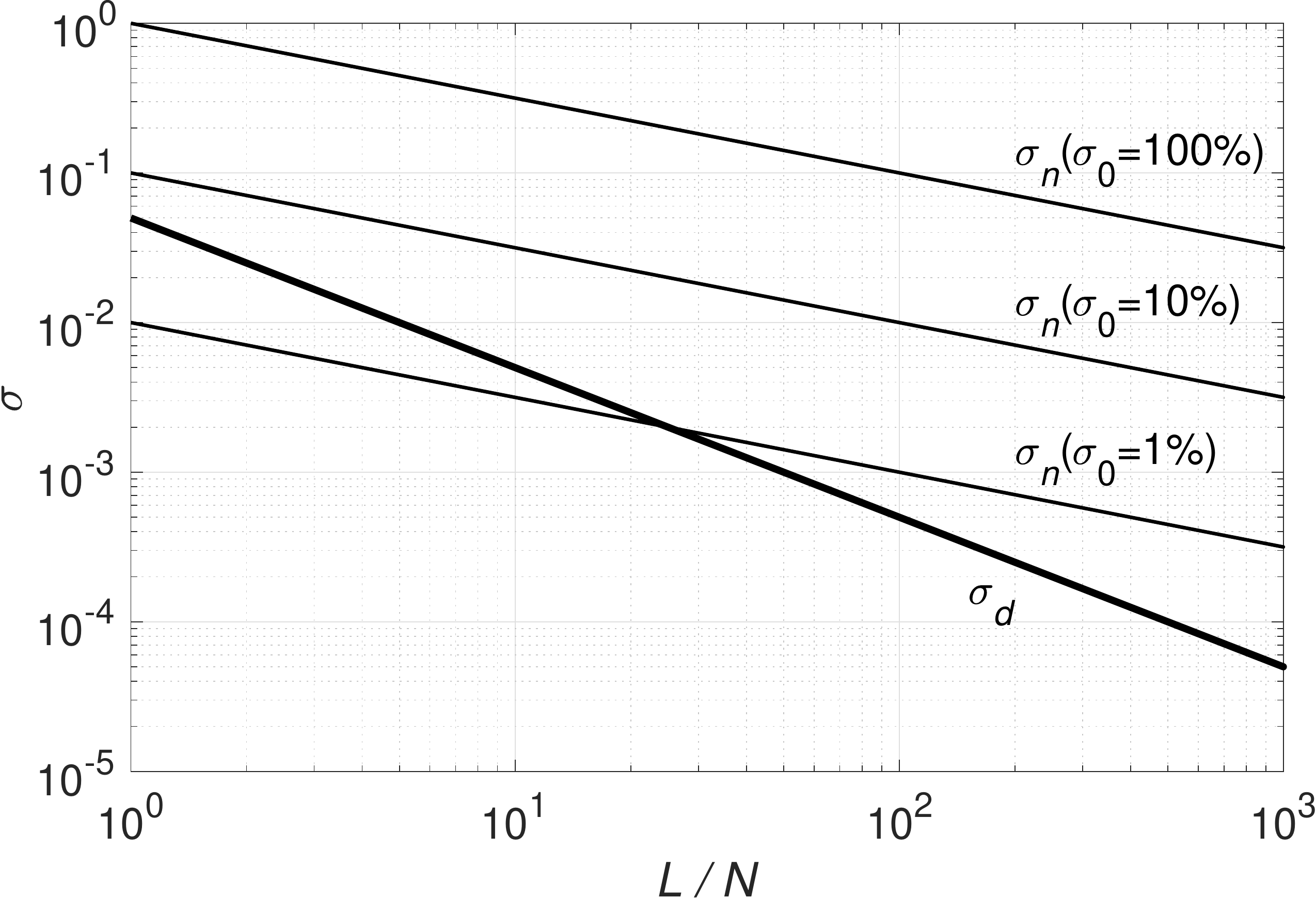}
		\caption{Uncertainty of the VIC method, for $R=2$}
		\label{graph_STD_L}
	\end{center}
\end{figure}
Fig.~\ref{graph_STD_L} shows the uncertainty of the method, according to expressions of $\sigma_d$ (Eq.~\ref{eq:sigma_disc}) and $\sigma_n$ (Eq.~\ref{eq:sigma}). 
It shows in particular that the irreducible uncertainty $\sigma_d$, associated to discretization, can only be attained with low image noise $\sigma_0$ and large curve support $L/N$.
Of course this graph is valid only in absence of curve fitting error.\\
%

%---------------------------------------------------------------
\section{Validation and comparison of the VIC uncertainty}\label{sec:3}
%---------------------------------------------------------------

%---------------------------------------------------------------
\subsection{Validation of the proposed expressions}\label{sec:31}

In table~\ref{tab:precision} we compare predicted and measured uncertainties on various tests.
Cases C$_1$ to C$_4$ correspond to numerical tests onto a $401\times401$ pixels image of a spiral \cite{francois11epj}.
Cases D$_1$ to D$_3$ refer to synthetic images of discs of average radii respectively of 3 (Fig.~\ref{fig:fff}), 10, 100 pixels whose center and radius are randomly varied over 1 pixel, over 100 trials.
Cases D$_1'$ to D$_3'$ are similar, but with an additive gaussian image noise $\sigma_0$.
All images are in 8 bits.
The standard deviation associated to discretization $\sigma_d$ is obtain by Eq.~(\ref{eq:sigma_disc}) and the one associated to image noise $\sigma_n$ by Eq.~(\ref{eq:sigma}).
The quantification noise $\sigma_{0q}$ is took into account in $\sigma_n$.
\begin{table}[h!]
\caption{Predicted ($\sigma_d$, $\sigma_n$) and measured ($\sigma$) standard deviations. Mesured mean ($m$)}
\begin{center}
\begin{tabular}{|r||r|r|r|r||r|r||r|r|}
\hline
case & $R$ & $L$ & $N$ & $\sigma_0$ & $\sigma_d\times 10^{3}$ & $\sigma_n\times 10^{3}$ & $\sigma\times 10^{3}$ &  $m\times 10^{3}$ \\
\hline
C$_1$ & $1$ & $1236$ & $20$ & $0\%$ & $0.809$ & $0.20$ & $9$ & \\
C$_2$ & $1$ & $1236$ & $20$ & $30\%$ & $0.809$ & $54.2$ & $54$ & \\
C$_3$ & $1$ & $1236$ & $20$ & $50\%$ & $0.809$ & $90.1$ & $85$ & \\
C$_4$ & $1$ & $1236$ & $20$ & $90\%$ & $0.809$ & $163$ & $180$ & \\
\hline
D$_1$ & $2$ & $18.8$ & $3$ & $0\%$ & $7.96$ & $8.41$ & $4$ & $-1.71$\\
D$_1'$ & $2$ & $18.8$ & $3$ & $10\%$ & $7.96$ & $48.3$ & $56$ & $2.12$\\
D$_2$ & $2$ & $62.8$ & $3$ & $0\%$ & $2.39$ & $2.63$ & $0.96$ & $-0.21$\\
D$_2'$ & $2$ & $62.8$ & $3$ & $10\%$ & $2.39$ & $24.5$ & $30$ & $0.26$\\
D$_3$ & $2$ & $628$ & $3$ & $0\%$ & $0.24$ & $0.32$ & $0.24$ & $0.00$\\
D$_3'$ & $2$ & $628$ & $3$ & $10\%$ & $0.24$ & $7.23$ & $8$ & $0.86$\\
%C$_1$ & $1$ & $1236$ & $20$ & $0\%$ & $0.809$ & $0.20$ & $9$ & \\
%C$_2$ & $1$ & $1236$ & $20$ & $30\%$ & $0.809$ & $54.2$ & $54$ & \\
%C$_3$ & $1$ & $1236$ & $20$ & $50\%$ & $0.809$ & $90.1$ & $85$ & \\
%C$_4$ & $1$ & $1236$ & $20$ & $90\%$ & $0.809$ & $163$ & $180$ & \\
%\hline
%D$_1$ & $2$ & $18.8$ & $3$ & $0\%$ & $7.96$ & $8.41$ & $4$ & $-1.71$\\
%D$_1'$ & $2$ & $18.8$ & $3$ & $10\%$ & $7.96$ & $48.3$ & $56$ & $2.12$\\
%D$_2$ & $2$ & $62.8$ & $3$ & $0\%$ & $2.39$ & $2.63$ & $0.96$ & $-0.21$\\
%D$_2'$ & $2$ & $62.8$ & $3$ & $10\%$ & $2.39$ & $24.5$ & $30$ & $0.26$\\
%D$_3$ & $2$ & $628$ & $3$ & $0\%$ & $0.24$ & $0.32$ & $0.24$ & $0.00$\\
%D$_3'$ & $2$ & $628$ & $3$ & $10\%$ & $0.24$ & $7.23$ & $8$ & $0.86$\\
\hline
\end{tabular}
\end{center}
\label{tab:precision}
\end{table}
One observes that the predicted values of $\mathrm{max}(\sigma_d,\sigma_n)$ are in good agreement with measured ones $\sigma$. The sole exception is the case C$_1$ for which a curve fitting error is present, the 10 control points of the B-Spline being not enough to describe the spiral at this level of precision.

%---------------------------------------------------------------
\subsection{Comparison between the VIC and other methods uncertainties}\label{sec:comparison}

The VIC method has been already successfully compared to Fast Marching Algorithm \cite{Sethian_98} and Steger's method \cite{steger1998} in earlier publication \cite{francois11epj}.
Since this article, new methods also claimed for sub-pixel precision.
Among them we retained the work of Trujilo-Pino (TP) \cite{Trujillo2013,Trujillo2019} which, based on an area estimate, is in some way close to the estimator $\mu$ (Eq.~\ref{eq:defmu}).
For reference, we retained the well known Active Contours (AC) method \cite{Lankton2019}.
\begin{table}[h!]
\caption{Measured uncertainty for active contours (AC) and Trujilo-Pino's (TP) methods}
\begin{center}
\begin{tabular}{|r||r|r||r|r|}
\hline
 & \multicolumn{2}{|c||}{AC} & \multicolumn{2}{c|}{TP} \\
\hline
case & $\sigma\times 10^{3}$ & $m\times 10^{3}$ & $\sigma\times 10^{3}$ & $m\times 10^{3}$ \\% $L$ & $\sigma_0$(\%) & 
\hline
D$_1$ & $44$ & $-93$ & $6$ & $3.94$\\ % $18.8$ & $0$ & 
D$_1'$ & $75$ & $-110$ & $227$ & $7.08$\\ % $18.8$ & $10$ & 
D$_2$ & $30$ & $-31$ & $2.04$ & $-0.12$\\ % $62.8$ & $0$ &
D$_2'$ & $71$ & $-35$ & $231$ & $-2.18$\\ % $62.8$ & $10$ &
D$_3$ & $27$ & $-4.52$ & $1.95$ & $0.01$\\ % $628$ & $0$ & 
D$_3'$ & $69$ & $-4.59$ & $236$ & $1.55$\\ % $628$ & $10$ &
\hline
\end{tabular}
\end{center}
\label{tab:pcomparison}
\end{table}
Table~\ref{tab:pcomparison} shows the results obtained for the circular disc statistical study.
With respect to TP method, the VIC (table~\ref{tab:precision}) offers a gain in $\sigma$ which increases with $L/N$.
In the realistic case D$_3$, with $L/N\simeq200$ pixels per curve parameter, the VIC is approximatively 6 times more precise than the TP method.
The AC method gives worse results but still identifies a continuous contour in noisy images, on the contrary of TP method which required to remove aberrants points (farther than 0.5 pixel).
Both TP and AC methods, like the majority of the existing contour detection methods, are local ones.
On the contrary, the VIC benefits of the regularization associated to the curve $\mathcal C$, whose effect on the precision, from Eqs.~(\ref{eq:sigma_disc}, \ref{eq:sigma}), increases with the curve length.

%---------------------------------------------------------------
\section{Conclusions}
%---------------------------------------------------------------

With a reliable expression of the uncertainty and a tool to estimate the relevance of the chosen curve family, the Virtual Image Correlation method has now reached maturity.
This article gives it a clarified theoretical framework and the sole parameter of the method, the virtual image width, is now fixed.\\

Relative interests between local and global methods are subjects of endless debates in the DIC community \cite{WANG2016200}. 
As expected, the VIC has the same advantages as the global DIC: accuracy and robustness to noise, but also shares its disadvantages: the necessity to choose an \emph{a priori} field (DIC) or curve (VIC).
Furthermore the given measure does not consist in a set of pixels but in a continuous curve defined from a reduced set of shape parameters.
The initialization step required for the VIC can be helped by temporarily setting a wide virtual image width $R$ or by using one of the many existing detection method, for example the robust Active Contours method.
Remaining possible ameliorations of the VIC may consist in faster computational strategies and some work remain to be done in 3D.\\

The field of applications is wide, especially in experimental mechanics.
The VIC can be used to measure object boundaries \cite{CONFSEM15,Jiang2015}, the shape of elongated objects (beam, trusses\dots) \cite{bloch2015} and possibly compare these curves between free and strained states.
The line of interest can also be a 2D \cite{CONFRILEM16} or 3D \cite{francois13exm} crack or a physical front (chemical, thermal, hydric\dots) \cite{francois10amm}.
Recent developments concern the use of the VIC to improve the DIC's precision close to the object borders \cite{baconnais19mdpi}.

\newpage
\appendix

%-----------------------------------------------------------------------------
\section{Relative magnitude of major terms}\label{sec:annex:magnitude}
%---------------------------------------------------------------

The relative magnitude of the terms in Eq.~(\ref{eq:d2psidl2}) are compared together in order to justify the use of the simplified Eq.~(\ref{eq:d2psidl2s}).
%The relative magnitude of the terms in \ref{eq:dpsidlp} and \ref{eq:d2psidl2} is compared below.
%In particular this shows the availability of the passage from \ref{eq:d2psidl2} to the simplified key equation \ref{eq:d2psidl2s}.
We suppose that, close to solution, $F(\mathbf{X}) \simeq f(x_2)$.
% the physical image depends only upon the distance $Rx_2$ to the curve (Eq.~(\ref{hyp:fx2})). 
If $\rho\ll R$, Eq.~(\ref{eq:dfdX}) can be expressed as:
% (an extension of the non-overlap condition Eq.~(\ref{cond:nonoverlap})):
\bq
%	\frac{\partial F}{\partial \vec X} &=& \frac{f'(x_2)}{R}\vec{e}_r\\
%	\frac{\partial^2 F}{\partial \vec X^2} &=& \frac{1}{R}\frac{\partial (f'\vec e_r)}{\partial\vec X}\\
%	\frac{\partial \vec{e}_r}{\partial \vec X} &=& \frac{\rho}{1+\rho R x_2} \vec e_s \otimes \vec e_s \label{eq:derdX}\\% Vérifiée
%	\frac{\partial f'(x_2)}{\partial \vec X} &=& \frac{f''(x_2)}{R}\vec e_r\\% OK
%
%	\frac{\partial^2 F}{\partial \vec X^2} &=&
%		\frac{f''(x_2)}{R^2} \vec{e}_r\otimes\vec{e}_r 
%		+
%		\frac{\rho f'(x_2)}{R(1+\rho R x_2)} \vec{e}_s\otimes\vec{e}_s
%	\label{eq:d2fdX2}\\% Vérifiée
	\frac{\partial^2 F}{\partial \vec X^2} &\simeq& \frac{f''(x_2)}{R^2} \vec{e}_r\otimes\vec{e}_r
\eq
where $\otimes$ denotes the dyadic (tensor) product.
Then, the terms of interest in Eqs.~(\ref{eq:dpsidlp}, \ref{eq:d2psidl2}) can be rewritten in a separate form:
\bq
%	I_1 &=& \int_{-1}^1\int_0^1
%	\left( \frac{\partial F}{\partial \vec X} \cdot \frac{\partial \vec X}{\partial \l q} \right)(f-g)dx_1 dx_2\nonumber\\
%	&\simeq& 
%	\frac{1}{R}\int_0^1 \vec e_r\cdot \frac{\partial \vec X^c}{\partial \l p} dx_1\quad
%	\int_{-1}^1 f'(f-g)dx_2\\
	I_2 &=& \int_{-1}^1\int_0^1
	\left(
	\frac{\partial \vec X}{\partial \l p}\cdot \frac{\partial^2 F}{\partial \vec X^2} \cdot \frac{\partial \vec X}{\partial \l q}
	\right)
	(f- g) dx_1 dx_2\nonumber\\
	&\simeq&
	\frac{1}{R^2}
	\int_0^1
	\left(\frac{\partial \vec X^c}{\partial \l p}\cdot\vec{e}_r\right)
	\left(\frac{\partial \vec X^c}{\partial \l q}\cdot\vec{e}_r\right)
	\,dx_1\quad
	\int_{-1}^1f''(f-g)dx_2
	\label{eq:I2}\\
	I_3 &=& \int_{-1}^1\int_0^1
	\left(
	\frac{\partial F}{\partial \vec X} \cdot \frac{\partial^2 \vec X}{\partial \l p\partial \l q}
	\right)
	(f- g) dx_1 dx_2\nonumber\\
	&\simeq&
%	\frac{1}{R}
%	\int_0^1
%	\left(\frac{\partial^2 \vec X^c}{\partial \l p\partial \l q}\cdot\vec{e}_r\right)\,dx_1
%	\int_{-1}^1f'(f-g)\,dx_2\nonumber\\
%	&+&
	\int_0^1
%	\left\| \frac{\partial \vec X^c}{\partial x_1} \right\|^2 
%	\left(\frac{\partial\vec e_s}{\partial \l p}\cdot\vec{e}_r\right)
%	\left(\frac{\partial\vec e_s}{\partial \l q}\cdot\vec{e}_r\right)
	\left(\frac{\partial^2 \vec X^c}{\partial \l p\partial x_1}\cdot\vec{e}_r\right)
	\left(\frac{\partial^2 \vec X^c}{\partial \l q\partial x_1}\cdot\vec{e}_r\right)
	d x_1\quad
	\int_{-1}^1 x_2f'(f-g)\,dx_2
	\label{eq:I3}\\
	I_4 &=&  \int_{-1}^1\int_0^1
	\left( \frac{\partial F}{\partial \vec X} \cdot \frac{\partial \vec X}{\partial \l p} \right)
	\left( \frac{\partial F}{\partial \vec X} \cdot \frac{\partial \vec X}{\partial \l q} \right)
	dx_1 dx_2\nonumber\\
	&=&
	\frac{1}{R^2}
	\int_0^1
	\left(\frac{\partial \vec X^c}{\partial \l p}\cdot\vec{e}_r\right)
	\left(\frac{\partial \vec X^c}{\partial \l q}\cdot\vec{e}_r\right)
	dx_1\quad
	\int_{-1}^1 (f')^2\,dx_2
	\label{eq:I4}
\eq
\begin{figure}[h!]
	\begin{center}
		\includegraphics[scale=0.50]{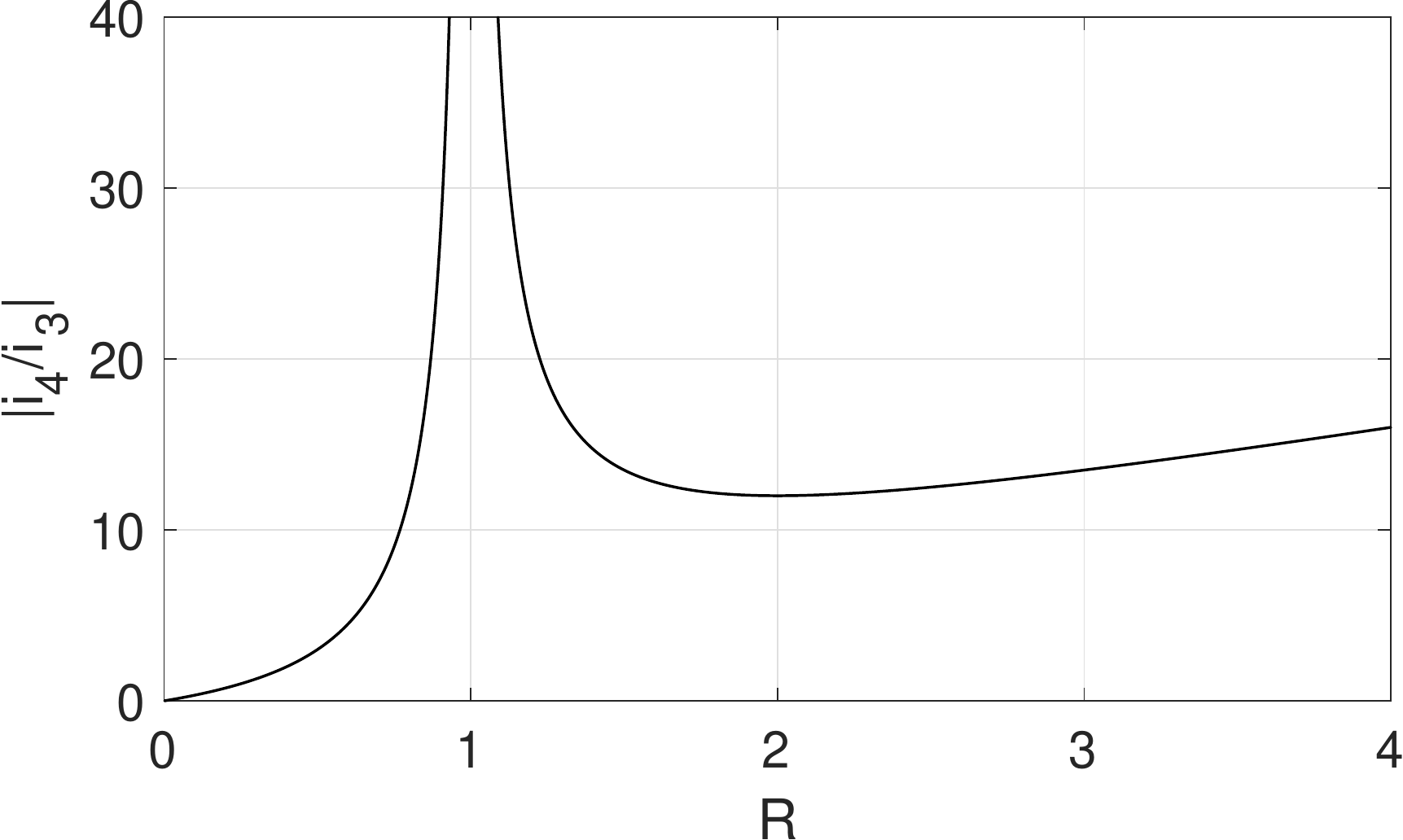}
		\caption{Ratio $|i_4/i_3|$}
		\label{fig:hyperbol}
	\end{center}
\end{figure}
Integrals over $x_2$ $(i_2,i_3,i_4)$, in correspondance with $(I_2,I_3,I_4)$, depend upon $g(x_2)$ (Eq.~\ref{eq:gdexsil}) and $f(x_2)$.
According to Eq.~(\ref{eq:linf}), in an ideal case $a=1$, $b=0$ and $\delta=0$, $f(x_2) = (1+Rx_2)/2$ thus:
$i_2=0$, 
$i_3=(R-1)/6R^2$ if $R>1$
or
$i_3=R(R-1)/6$ if $R<1$
and
$i_4=1/2$ if $R>1$ or
$i_4=R^2/2$ else.
%
%$f(x_2)={(\alpha+x_2)}/{2\alpha}$.
%$i_2=0$, 
%$i_3=\alpha(1-\alpha)/6$ if $\alpha<1$
%or
%$i_3=(1-\alpha)/6\alpha^2$ if $\alpha>1$
%and
%$i_4=1/2$ if $\alpha<1$ or
%$i_4=1/2\alpha^2$ else.
%
Fig.~\ref{fig:hyperbol} shows that $i_3\ll i_4$ as soon as $R>1$, \emph{i.e} as soon as the virtual image width is wide enough to cover the black to white transition in F. 
% For the value $R=1.5$ pixel retained in Sec.~\ref{sec:precimes}, this ratio is $13.5$. 
This result justifies the simplification from Eqs.~(\ref{eq:d2psidl2}, \ref{eq:d2psidl2s}), as soon as the integrals over $x_1$ have comparable magnitudes.
%The function $f(x_2)$ is practically obtained from a linear approximation of the digital image F over the sub-pixel coordinates (Fig.~\ref{fig:circle}, right) thus is close to a continuous segment curve (Fig.~\ref{fig:1D_vision}) which is approximated here by a single linear function of equation: 
%\bq
%	\frac{\partial \vec{e}_s}{\partial x_1} &=& 
%	- \rho \left\| \frac{\partial \vec X^c}{\partial x_1} \right\|\vec{e}_r \label{desdx1}\label{derdx1}\\
%	\frac{\partial \vec{e}_s}{\partial \l p}
%	&=&
%	\left\| \frac{\partial \vec X^c}{\partial x_1} \right\|^{-1}
%	\left(\frac{\partial^2\vec X^c}{\partial \l p\partial x_1}\cdot\vec e_r\right) \vec e_r \label{eq:desdlp}
%\eq	

%-----------------------------------------------------------------------------
\section{Intermediate calculations on the effet of image noise}\label{sec:annex:noise}
%---------------------------------------------------------------
%
Hypotheses of Sec.~\ref{sec:idcases}, Eqs.~(\ref{eq:dpsidlp_fx2}) and (\ref{eq:vicint}) lead to:
\be
	\frac{\partial \psi}{\partial \l p} 
	=
	\frac{1}{2R}
	\int_0^1 \int_{-1}^1
	\frac{\partial \vec X^c}{\partial \l p}
	\cdot
	\vec{e}_r\,
	 \left( f(x_2)-\frac{1}{2} \right) dx_1 dx_2 = 0\label{eq:dpsidlp1}
\ee
%
%.In a numerical point of view this expression is computed in the pixel coordinates $\vec X$.
Eqs.~(\ref{eq:revmap} to \ref{def:rho}) give the differential surface element in the frame $\vec X$:
\be
	d\vec X = \left( \frac{\partial \vec X^c}{\partial x_1}+\rho R x_2\left\| \frac{\partial \vec X^c}{\partial x_1} \right\| \vec e_s \right) dx_1+Rdx_2\,\vec e_r\label{eq:dXdx}
\ee
Supposing a weak curvature $|\rho| R\ll 1$ we obtain:
\be
	dX_1dX_2 \simeq RL dx_1 dx_2
\ee
which gives the correspondance between the virtual image surfaces $S=2RL$ in the frame $\vec X$ and $s=2$ in the frame $\vec x$.
% Even if the computation is proceeded in a refined frame issued from the discretization of ($x_1,x_2)$, the data refer to the pixel frame associated to the pixel coordinates corresponding to the integer values of $\vec X$.
In the pixel frame, Eq.~(\ref{eq:dpsidlp1}) corresponds to:
\be
	\frac{\partial \psi}{\partial \l p} \simeq \frac{1}{Rn}\sum_{ij} 
	\left( \frac{\partial \vec X^c}{\partial \l p} \cdot \vec{e}_r \right)_{ij}
	\left( F_{ij}+N_{ij}-\frac{1}{2} \right)
\ee
where $n$ is the number of pixels involved in the virtual image calculus thus $n\simeq 2RL$ (the virtual image surface).
%One may note that the effect of the sub-pixel grid (Section~\ref{sec:discretization}) is not taken into account because it does not increase the number of physical data (the pixels). 
%Supposing the pixel index $i$ (mainly) related to the curvilinear abscissa and index $j$ related to the transverse direction
From elementary statistics, we obtain the standard deviation:
\be
	\sigma\left(\frac{\partial \psi}{\partial \l p}\right) \simeq \frac{\sigma_0}{2LR^2}
	\sqrt{\sum_{ij} \left( \frac{\partial \vec X^c}{\partial \l p} \cdot \vec{e}_r \right)_{ij}^2}
\ee
and Eq.~(\ref{eq:sigmadpsidlp}) is deduced from this equation and the following correspondance between continuous and discrete expressions:
\be
	\sum_{ij}\left( \frac{\partial \vec X^c}{\partial \l p}\right)^2 \simeq 2RL\int_0^1 \left( \frac{\partial \vec X^c}{\partial \l p}\right)^2 dx_1,
\ee
%

%--------------------------------------------------------------
%\begin{acknowledgements}
%\end{acknowledgements}


\bibliographystyle{plain}
\bibliography{reference}   % name your BibTeX data base

%\begin{thebibliography}{}
% etc
%\end{thebibliography}

\end{document}